%% file: acl_latex.tex
\documentclass[11pt]{article}

\usepackage[final]{acl}

\usepackage{times}
\usepackage{latexsym}
\usepackage{stfloats} 
\usepackage[T1]{fontenc}

\usepackage[utf8]{inputenc}

\usepackage{microtype}

\usepackage{inconsolata}

\usepackage{graphicx}
\usepackage{amsfonts}
\usepackage{amssymb}
\usepackage{xcolor}    
\usepackage[table]{xcolor}
\definecolor{LightCyan}{RGB}{224,255,255}
\usepackage{hyperref}  
\usepackage{url}
\usepackage{multirow}
\usepackage{titletoc}
\usepackage{tocbibind} 
\usepackage{hyperref}  
\usepackage{etoolbox}  
\usepackage{graphicx}
\usepackage{booktabs}
\usepackage{enumitem}
\usepackage{epigraph}
\usepackage{makecell}
\usepackage{tabularray}
\usepackage{algorithm}
\usepackage{algpseudocode}
\usepackage{wrapfig}
\usepackage{pifont}
\usepackage{algpseudocode} 
\usepackage{tikz}
\usepackage{amsmath}
\usepackage[table]{xcolor}
\usepackage{tabularray}
\UseTblrLibrary{siunitx,booktabs}
\usepackage{siunitx}
\usepackage{siunitx}
\definecolor{selfevolagent_blue}{HTML}{0064E0}
\hypersetup{
  colorlinks=true,
  citecolor=selfevolagent_blue,
  urlcolor=selfevolagent_blue,
  linkcolor=blue
}

\definecolor{darkred}{RGB}{170,70,70}           
\definecolor{forestgreen}{RGB}{90,140,90}     
\definecolor{QwenPurple}{HTML}{5B4BE6}
\newcommand{\down}[1]{\textcolor{darkred}{\scriptsize \rlap{$\,\downarrow$}#1}}
\newcommand{\up}[1]{\textcolor{forestgreen}{\scriptsize \rlap{$\,\uparrow$}#1}}

\newcommand{\github}{\raisebox{-2pt}{\includegraphics[height=1.05em]{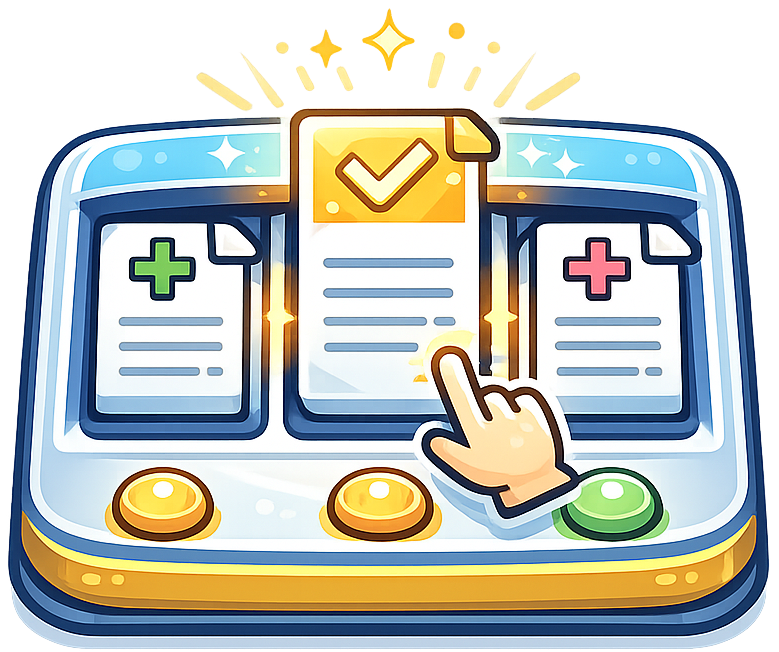}}}
\newcommand{\ghlink}{https://x-izhang.github.io/CCS/}
\title{\textbf{CCS}: \underline{C}linical \underline{C}onsensus \underline{S}election for Radiology Report Generation}

\author{Xi Zhang$~^{\spadesuit}$, Yingshu Li$~^{\clubsuit}$, Zaiqiao Meng$~^{\spadesuit,\diamondsuit}$, Jake Lever$~^{\spadesuit}$, Edmond S. L. Ho$~^{\spadesuit}$\\
  $^{\spadesuit}$School of Computing Science, University of Glasgow\\
  $^{\clubsuit}$School of Electrical and Computer Engineering, University of Sydney\\
  $^{\diamondsuit}$Language Technology Lab, University of Cambridge\\
  \texttt{X.Zhang.6@research.gla.ac.uk}\\
  \texttt{yingshu.li@sydney.edu.au, mz468@cam.ac.uk}\\
  \texttt{Jake.Lever@glasgow.ac.uk, Shu-Lim.Ho@glasgow.ac.uk}\\
  \makebox[0pt][c]{%
  \raisebox{-0.2em}{%
    \begin{tabular}{c}
        \github\hspace{0.1em}      \url{\ghlink}
    \end{tabular}
    }
  }
}

\begin{document}
\setlength{\abovedisplayskip}{4pt}
\setlength{\belowdisplayskip}{4pt}
\setlength{\abovedisplayshortskip}{2pt}
\setlength{\belowdisplayshortskip}{2pt}
\maketitle
\begin{abstract}
Radiology report generation (RRG) is commonly formulated as a single-path generation task, where a multimodal large language model (MLLM) produces one decoded report as the final output. While recent progress has largely been driven by scaling training data, model capacity, and retrieval mechanisms, improving report quality at inference time remains underexplored. 
In this work, we observe that fixed radiology MLLMs often generate clinically stronger reports elsewhere in their candidate pool than the one selected by default decoding, suggesting that inference-time decision making remains an overlooked bottleneck. 
To address this, we propose \underline{\textbf{C}}linical \underline{\textbf{C}}onsensus \underline{\textbf{S}}election (\textbf{\textsc{CCS}}), a \textit{decoder-agnostic} inference-time selection framework that samples multiple candidate reports and selects the one with the highest clinical consensus across the rollout pool. 
\textsc{CCS} unifies text-based utilities with a radiology-adapted utility computed by an image--report-trained multimodal embedder, which measures candidate agreement beyond surface-level textual similarity.
Across three datasets and multiple radiology MLLMs, \textsc{CCS} consistently improves inference-time performance over single-path decoding and generic Best-of-$N$ baselines, with particularly clear gains on clinical metrics. Further analysis shows that image-grounded utility forms a selection axis distinct from textual consensus and that substantial headroom remains for improving RRG at inference time.
\end{abstract}

\input{sections/intro}

\input{sections/related}

\input{sections/method}

\input{sections/experiment}

\input{sections/results}

\input{sections/conclusion}

\clearpage

\input{sections/limitations}

\input{sections/ethical}

\bibliography{custom}

\newpage
\appendix


\section*{Appendix Contents}
\startcontents[sections]
\printcontents[sections]{l}{1}{%
    \setcounter{tocdepth}{2}
}

\newpage

\input{sections/appendix}

\end{document}

%% file: sections/intro.tex
\section{Introduction}
\label{sec:intro}

\begin{figure}[t]
    \centering
    \includegraphics[width=\columnwidth]{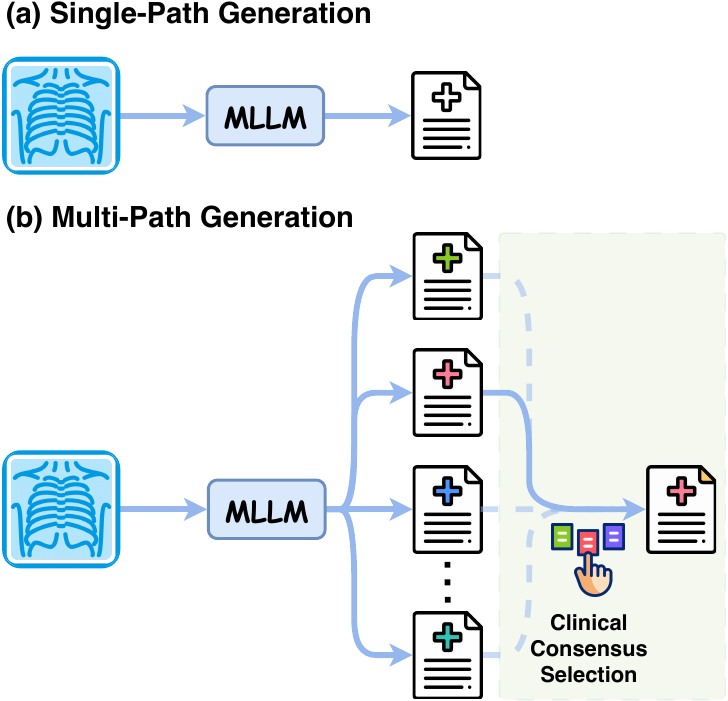}
    \vspace{-20pt}
    \caption{
    \textbf{From Single-Path Generation to Clinical Consensus Selection (CCS).} 
    \textbf{(a)} Conventional RRG systems ultimately return one decoded report as the final output; 
    \textbf{(b)} CCS forms a candidate rollout pool and selects the report with higher relative clinical consensus.
    }
    \label{fig:case-problem}
    \vspace{-10pt}
\end{figure}

Radiology report generation (RRG) aims to express clinical findings from radiology images, such as chest X-rays, as free-text reports, forming a core component of the radiology workflow~\citep{liu2019clinicallyaccuratechestxray,monshi2020deep}. Recent multimodal large language models (MLLMs) have driven substantial progress on this task by scaling model capacity~\citep{tu2023generalistbiomedicalai,li2023llavamed}, training data~\citep{bannur2024maira2groundedradiologyreport,Zambrano_Chaves_2025}, and retrieval-augmented generation~\citep{xia2025mmedragversatilemultimodalrag,hou2025radarenhancingradiologyreport}. However, comparatively less attention has been paid to improving report quality \emph{at inference time}, where the model parameters and external evidence are fixed.

Despite this progress, automated chest X-ray report generation remains far from meeting the demands of real-world clinical practice~\citep{zhang2025automated}. Most MLLMs still rely on single-path generation, committing to one report token by token (Figure~\ref{fig:case-problem}-\textbf{a}), and even recent test-time refinements, such as clinical contrastive decoding~\citep{zhang2025ccd}, follow a single decoded trajectory. This is fragile: one unfavourable decoding step can omit a finding or assert one unsupported by the image, with no mechanism for recovery. 
In this work, we observe that a fixed model often places clinically stronger reports elsewhere in its candidate pool than the one returned by default decoding, leaving a gap to the pool-bounded oracle (as shown in Figure~\ref{fig:metric_trend_sampling}).
The bottleneck lies not in what the model can generate, but in which candidate it commits to, suggesting that inference-time decision making is an underexplored opportunity for improving RRG without modifying or retraining the model.

Selecting among multiple generations has become a key mechanism for improving generation quality at test time, as seen in Best-of-\textit{N}~\citep{snell2024scaling,hu2024can,huang2025best} and self-consistency methods~\citep{wang2024soft,kang2026scalable,choi2026modex}. However, existing selection criteria are not designed for radiology reports. Fluency, average log-probability, and textual agreement may favour plausible-sounding or conservative reports, but clinical correctness cannot be reduced to surface quality, token confidence, or text-only similarity. This is especially problematic for open-ended RRG, where multiple phrasings can be clinically equivalent and no reference report is available at test time. Effective inference-time optimisation therefore requires identifying candidates with high clinical consensus in radiology-adapted representation spaces, rather than relying only on conventional text-based signals (e.g., perplexity).

To address this, we propose \underline{\textbf{C}}linical \underline{\textbf{C}}onsensus \underline{\textbf{S}}election (\textbf{\textsc{CCS}}), a \textit{decoder-agnostic} inference-time selection framework for RRG (Figure~\ref{fig:case-problem}-\textbf{b}).
Given a rollout pool from a radiology MLLM, \textsc{CCS} scores candidate pairs with a pluggable utility and returns the report with the highest mean consensus over the pool. 
We instantiate a radiology-adapted utility using Qwen3-VL-Embed~\citep{li2026qwen3}, a multimodal embedder adapted on image--report pairs, which measures candidate agreement in a radiology representation space and provides a signal beyond text-only similarity, particularly for symptom-level findings. Our contributions are:

\begin{itemize}[align=right,itemindent=2em,labelsep=2pt,labelwidth=1em,leftmargin=0pt,nosep,topsep=4pt,label=$\circ$]
    \setlength{\itemsep}{4pt}
    \item[\ding{182}] We revisit the RRG task from an \textbf{inference-time perspective} and show that candidate pools routinely contain reports with higher clinical reliability and consistency than single-path outputs.
    \item[\ding{183}] We propose \textbf{\textsc{CCS}}, a decoder-agnostic Best-of-\textit{N} framework that aggregates pairwise clinical consensus over a candidate pool using textual and image--report-adapted multimodal utilities.
    \item[\ding{184}] Extensive experiments across three datasets, multiple radiology MLLMs, and qualitative case analyses show that \textbf{\textsc{CCS}} consistently improves backbone performance for RRG at inference time, while identifying image-grounded utility as a distinct selection axis beyond textual consensus.
\end{itemize}

%% file: sections/related.tex
\newpage
\section{Related Work}
\label{sec:related}

\subsection{Radiology Report Generation}

RRG aims to generate clinically coherent reports from medical images. Early methods typically adopt encoder--decoder architectures trained on paired image--report data~\citep{liu2019clinicallyaccuratechestxray,monshi2020deep,wang2018tienettextimageembeddingnetwork}. Recent work extends this paradigm with radiology MLLMs, including LLaVA-Med~\citep{li2023llavamed}, LLaVA-Rad~\citep{Zambrano_Chaves_2025}, Libra~\citep{zhang2025libraleveragingtemporalimages}, MAIRA~\citep{hyland2024maira1specialisedlargemultimodal,bannur2024maira2groundedradiologyreport}, and biomedical foundation models~\citep{tu2023generalistbiomedicalai}, often further enhanced by retrieval augmentation~\citep{sun2025factawaremultimodalretrievalaugmentation}.

However, most RRG methods still follow a single-trajectory inference paradigm. While token-level methods such as contrastive decoding or logit manipulation~\citep{li2023contrastivedecodingopenendedtext,zhang2025ccd} adjust generation locally, \textsc{CCS} further optimises inference through reference-free candidate selection.

\subsection{Inference-Time Optimisation}

Inference-time optimisation improves generation by allocating extra decoding-time computation without updating model parameters~\citep{snell2024scaling,huang2025best}. Common strategies include Best-of-\textit{N} reranking, self-consistency~\citep{wang2024soft}, rollout-based selection~\citep{shao2024deepseekmath}, and reference-free scoring via likelihood, confidence, or text agreement~\citep{hu2024can,kang2026scalable,choi2026modex}.

However, scoring criteria based on likelihood, confidence, or text agreement are poorly suited to RRG, where lexically similar reports may differ in findings, anatomy, laterality, or temporal interpretation. \textsc{CCS} instead selects the report with the highest clinical consensus within the rollout pool.

\subsection{Multimodal Embeddings}

Multimodal embedding models learn shared representations across images and text, ranging from general-domain contrastive models~\citep{radford2021learningtransferablevisualmodels,zhai2023sigmoidlosslanguageimage} to instruction-tuned embedders~\citep{meng2025vlm2vecv2advancingmultimodalembedding} and biomedical variants~\citep{zhang2025biomedclipmultimodalbiomedicalfoundation,perezgarcia2024raddino}. These models are primarily developed for retrieval or representation learning rather than report selection. \textsc{CCS} repurposes radiology-adapted multimodal embeddings~\citep{li2026qwen3} as utility functions for candidate comparison, enabling image-grounded consensus estimation during inference.

%% file: sections/method.tex
\section{Clinical Consensus Selection}
\label{sec:methodology}

\paragraph{Rethinking Radiology Report Generation.}
A key challenge in inference-time RRG is that report quality cannot be directly verified. Rollout-based methods in reasoning LLMs, such as Group Relative Policy Optimisation (GRPO)~\citep{shao2024deepseekmath}, improve outputs by sampling multiple trajectories and exploiting relative reward signals. However, these approaches typically assume \emph{verifiable rewards}, such as mathematical correctness or executable code outcomes. RRG violates this assumption. At test time, the ground-truth report is unavailable, and no rule-based checker can determine whether a generated report is clinically correct. Moreover, clinical quality cannot be reduced to lexical or semantic similarity: reports with similar surface forms may differ substantially in findings, anatomy, laterality, or temporal interpretation. This motivates a central question: \emph{Can we select a clinically coherent report from multiple generations without access to any reference report?}

\begin{figure*}[t]
    \centering
    \includegraphics[width=\textwidth,keepaspectratio]{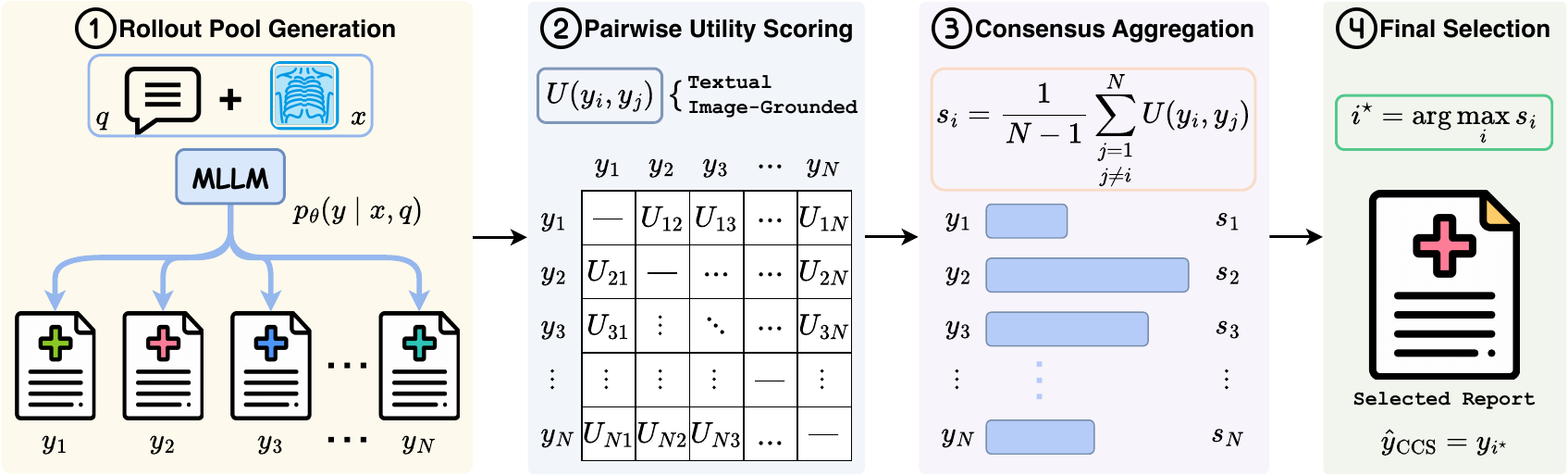}
    \vspace{-20pt}
    \caption{
    \textbf{Overview of the Clinical Consensus Selection framework.}
    At inference time, CCS proceeds in four stages:
    \textbf{(1)} constructing a rollout pool from a radiology MLLM;
    \textbf{(2)} computing pairwise utilities among candidates;
    \textbf{(3)} aggregating them into relative consensus scores;
    and \textbf{(4)} selecting the final report according to relative consensus.
    }
    \label{fig:framework}
    \vspace{-5pt}
\end{figure*}

We address this question through \textbf{Clinical Consensus Selection (CCS)}, a reference-free inference-time framework for RRG. Instead of returning the first decoded output, CCS samples a rollout pool and selects the final report according to clinical consensus among candidate generations (Figure~\ref{fig:framework}).


\subsection{Problem Formulation}
RRG is conventionally formulated as conditional sequence generation. Given a chest X-ray $x$ and a question $q$, a radiology MLLM parameterised by $\theta$ defines a distribution $p_\theta(y \mid x,q)$ over free-text reports $y$. The single-path paradigm returns one decoded report as the final output:
\begin{equation}
\hat{y}_{\text{single}} \sim p_\theta(y \mid x,q).
\label{eq:single}
\end{equation}
Since $\hat{y}_{\text{single}}$ is committed to one decoding trajectory, its clinical quality depends on one sampled or greedily selected sequence, without a mechanism to recover from omitted observations or unsupported findings. CCS instead reformulates inference as candidate selection over a rollout pool.

\subsection{Rollout Pool Generation}

The first stage constructs a rollout pool of candidate reports, from which the final report will be selected (stage \raisebox{.5pt}{\textcircled{\raisebox{-.9pt}{1}}} in Figure~\ref{fig:framework}). Given the same input $(x,q)$, we sample $N$ candidate reports from the MLLM under stochastic decoding with temperature $\tau$~\footnote{Unless otherwise specified, decoding hyperparameters such as top-$p$ and top-$k$ use Transformers library defaults.}:
\begin{equation}
\mathcal{Y}=\{y_1,\ldots,y_N\}, \quad
y_i \sim p_\theta(y \mid x,q;\tau).
\label{eq:pool}
\end{equation}
This stage leaves the generator unchanged: it introduces no additional parameters, retraining, or auxiliary supervision, and only varies stochastic decoding at inference time. The pool size $N$ and temperature $\tau$ determine the candidate space available to the downstream selector.

\subsection{Pairwise Utility Scoring}
\label{sec:method:utility}

The second stage measures pairwise agreement among candidates in the rollout pool. For each pair $(y_i,y_j)$, we compute a utility score $U(y_i,y_j)$ and form a score matrix $S\in\mathbb{R}^{N\times N}$, where $S_{ij}=U(y_i,y_j)$. 
We consider two utility families.

\paragraph{Textual Utility.}
These repurpose report evaluation metrics, detailed in \S\ref{subsec:metric}, as reference-free pairwise scores. Given a metric $m(\cdot,\cdot)$, we define
\begin{equation}
U_{\text{text}}(y_i,y_j)=m(y_i,y_j).
\label{eq:util_text}
\end{equation}
A higher score indicates stronger agreement between two generated reports under the chosen metric, yielding a metric-specific textual selector.

\paragraph{Image-Grounded Utility.}
Textual utilities compare reports without explicitly modelling whether agreement is grounded in the image. Inspired by universal multimodal embedding models~\citep{meng2025vlm2vecv2advancingmultimodalembedding}, we adapt Qwen3-VL-Embed~\citep{li2026qwen3} to the RRG task and use it as a report encoder $f_\phi$~\footnote{The embedder $f_\phi$ is adapted using image--report pairs for RRG, but the inference-time utility operates only over candidate reports and does not directly use the test image $x$.}. Given two candidates, we compute their similarity in the learned representation space:
\begin{equation}
U_{\text{img}}(y_i,y_j)=
\operatorname{CosineSim}\!\big(f_\phi(y_i),f_\phi(y_j)\big)
\label{eq:util_img}
\end{equation}
This utility favours candidate reports with high agreement in an RRG-adapted representation space, rather than surface-level textual overlap.

\subsection{Consensus Aggregation}
\label{sec:method:consensus}

The final stage aggregates pairwise scores into a consensus value for each candidate and returns the highest-scoring report (stages \raisebox{.5pt}{\textcircled{\raisebox{-.9pt}{3}}}--\raisebox{.5pt}{\textcircled{\raisebox{-.9pt}{4}}} in Figure~\ref{fig:framework}). 
Given a score matrix from any utility function, CCS applies the same aggregation rule across all selectors.
We score each candidate by its mean pairwise utility against other $N-1$ candidates in the pool,
\begin{equation}
s_i = \frac{1}{N-1}\sum_{\substack{j=1 \\ j\neq i}}^{N} U(y_i, y_j).
\label{eq:consensus}
\end{equation}
A high $s_i$ indicates that $y_i$ agrees with the pool under the chosen utility. CCS then selects the candidate with the highest consensus score:
\begin{equation}
\hat{y}_{\mathrm{CCS}} = y_{i^\star}, \quad
i^\star = \arg\max_{i\in\{1,\dots,N\}} s_i.
\label{eq:select}
\end{equation}

Algorithm~\ref{alg:ccs} summarises the overall CCS procedure, where the same aggregation rule is applied across selectors with different utility functions $U$.

\begin{algorithm}[t]
\caption{Clinical Consensus Selection}
\label{alg:ccs}
\small
\begin{algorithmic}[1]
\Require Test image $x$ and question $q$; radiology MLLM generator $p_\theta(y \mid x,q)$; pairwise utility function $U(\cdot,\cdot)$; pool size $N$; sampling temperature $\tau$
\Ensure Selected report $\hat{y}_{\mathrm{CCS}}$

\State Generate a rollout pool $\mathcal{Y}=\{y_1,\ldots,y_N\}$ by sampling from $p_\theta(\cdot \mid x,q)$ at temperature $\tau$
\Comment{candidate reports}

\For{$i = 1$ \textbf{to} $N$}
    \For{$j = 1$ \textbf{to} $N$}
        \State $S_{ij} \gets U(y_i,y_j)$ \Comment{pairwise utility}
    \EndFor
\EndFor

\State $\mathbf{s} \gets \dfrac{1}{N-1}\left(S\mathbf{1}-\mathrm{diag}(S)\right)$ 
\Comment{consensus utility}
\State $i^{\star} \gets \arg\max_{i\in\{1,\ldots,N\}} \mathbf{s}_i$
\State $\hat{y}_{\mathrm{CCS}} \gets y_{i^{\star}}$
\State \Return $\hat{y}_{\mathrm{CCS}}$
\end{algorithmic}
\end{algorithm}

%% file: sections/experiment.tex
\section{Experiments}
\label{sec:experiment}
\subsection{Datasets}
We evaluate our method on three publicly available radiology datasets: the official test splits of MIMIC-CXR~\citep{johnson2019mimic} and IU-Xray~\citep{demner2015preparing}, and the public validation set of CheXpert Plus~\citep{chambon2024chexpert}, as CheXpert Plus does not provide an official test split. Notably, all trainable models used in our experiments are trained only on the MIMIC-CXR training set, enabling us to assess cross-dataset generalisation on IU-Xray and CheXpert Plus without additional dataset-specific training.  Following prior work~\citep{Zambrano_Chaves_2025}, we focus on generating the \textit{findings} section from a single frontal-view image. Further details on dataset description and preprocessing are provided in Appx.~\ref{app:datasets}.

\subsection{Evaluation Metrics}
\label{subsec:metric}
Following prior research~\citep{hyland2024maira1specialisedlargemultimodal,hou2025radarenhancingradiologyreport}, we report standard lexical and radiology-specific RRG metrics. Lexical metrics, including ROUGE-L~\citep{lin-2004-rouge}, BLEU~\citep{10.3115/1073083.1073135}, and BERTScore~\citep{zhang2020bertscoreevaluatingtextgeneration}, assess textual similarity to reference reports. Radiology-specific metrics assess clinical correctness from complementary perspectives, including entity and relation overlap with RadGraph-F1~\citep{delbrouck-etal-2022-improving}, concept-level correctness with RaTEScore~\citep{zhao2024ratescoremetricradiologyreport}, semantic consistency with RadEval-BERT~\citep{xu-etal-2025-radeval}, and common finding coverage with CheXbert-F1~\citep{smit2020chexbertcombiningautomaticlabelers}. Detailed metric definitions and implementation details are provided in Appx.~\ref{app:metrics}.

\subsection{Baselines}
We compare \textsc{CCS} against the default \textit{Single-Path} generation setting, reporting both greedy and sampling-based decoding results. We also include three Best-of-\textit{N} selection baselines adapted from the general domain.
Perplexity selects the candidate with the lowest average uncertainty~\citep{hu2024can}; Self-Certainty~\citep{kang2026scalable} selects the candidate with the lowest negative log-likelihood; and ModeX~\citep{choi2026modex} constructs a text-similarity graph over candidate generations and selects the cluster centroid as the final output. As a sanity-check baseline, Random uniformly selects one candidate from the generated pool. To assess the generality of \textsc{CCS}, experiments are further conducted on several pre-trained radiology MLLMs, including LLaVA-Med~\citep{li2023llavamed}, LLaVA-Rad~\citep{Zambrano_Chaves_2025}, and Libra~\citep{zhang2025libraleveragingtemporalimages}. Additional details of these models are provided in Appx.~\ref{app:mllms}.

\subsection{Implementation Details}

\paragraph{Training.}
The baseline MLLM follows the LLaVA architecture~\citep{liu2023visualinstructiontuning}, consisting of a CLIP visual encoder~\citep{radford2021learningtransferablevisualmodels} and Vicuna-1.5~\citep{vicuna2023} as the language backbone. Following prior work~\citep{li2023llavamed}, training is conducted in two stages: Stage I trains only a two-layer MLP adapter for CXR--text feature alignment, while Stage II fine-tunes only the LoRA~\citep{hu2021loralowrankadaptationlarge} parameters of the LLM to improve RRG performance. In addition, Qwen3-VL-Embed-2B~\citep{li2026qwen3} is initialised from its pre-trained checkpoint and further adapted for CXR--report representation learning using the same training dataset as the baseline MLLM. Detailed training settings are provided in Appx.~\ref{app:train_details}.

\paragraph{Inference.}
For all evaluated methods, we follow the default inference configurations from their original papers where applicable. For MLLM-based report generation, the maximum generation length is set to 256 tokens; sampling-based decoding uses a temperature of $0.5$. Unless otherwise specified, Best-of-\textit{N} methods use a rollout pool of $N=8$ candidate reports; additional results with varying rollout sizes are reported in the analysis section. For Qwen3-VL-Embed, images are processed using the official Qwen-VL preprocessing pipeline. For reproducibility, the prompt templates used during training and inference are provided in Appx.~\ref{app:prompt}.

%% file: sections/results.tex
\section{Results and Analyses}

\subsection{Main Results}
\label{sec:main-results}

\input{tables/main_result_part_1}

\input{tables/other_MLLMs}

\paragraph{Comparison with Generic Best-of-\textit{N}.}

As shown in Table~\ref{tab:main_result_1}, generic Best-of-\textit{N} selectors yield limited and inconsistent gains over Sampling, reflecting distinct selection biases. Perplexity favours fluent candidates and improves lexical metrics, but brings limited clinical gains. ModeX, based on similarity-based clustering, provides moderate improvements yet remains below Sampling on $\mathrm{CheXbert}_{\mathrm{F1}}^{14}$. Self-Certainty underperforms across all metrics, suggesting that token-level confidence is poorly aligned with clinical correctness. Differences from Random selection further indicate that these gains mainly arise from utility-based selection rather than candidate re-sampling. 
In contrast, \textbf{\textsc{CCS}}, instantiated with Qwen3-VL-Embed utility, consistently improves performance across all metrics, with especially noticeable gains on radiology-specific metrics. 
Compared with Sampling, all observed improvements are statistically significant ($p<0.05$), based on paired approximate randomisation with $10{,}000$ random sign-flips; confidence intervals are computed using bootstrap resampling at the 95\% level. 
These findings suggest that rollout pools contain substantially better candidates than the first decoded output, and that \textsc{CCS} can identify them more effectively than generic approaches.

\paragraph{Cross-Backbone Consistency.}

We further examine the cross-backbone and cross-dataset behaviour of \textsc{CCS}. As shown in Table~\ref{tab:other_mllms}, \textsc{CCS} yields consistent clinical gains across all evaluated backbone--dataset settings. In particular, every radiology-specific metric improves over the corresponding Sampling baseline, suggesting that clinical consensus selection is not tied to a specific generator or data distribution. Lexical metrics occasionally decline, which is expected for a radiology-adapted utility that prioritises clinically meaningful agreement over surface overlap with common report phrasing. Overall, these results provide directional evidence that \textsc{CCS} can recover clinically stronger candidates across backbones and datasets.

\subsection{Consensus Utility Ablation}
\label{sec:utility-ablation}
\input{tables/main_result_part_2}

Table~\ref{tab:main_result_2} compares different consensus utilities on a shared rollout pool. A clear self-alignment pattern emerges: most utilities perform best on the metric from which they are derived, as consensus and evaluation rely on the same scoring signal. However, self-alignment does not necessarily translate to better symptom-label consensus. $\mathrm{CheXbert}$ metrics are dominated by frequent negative findings, making agreement on ``no finding'' cases easier than consensus on abnormal labels. As a result, label-based utilities may improve apparent label agreement without reliably identifying clinically meaningful abnormalities. By comparison, the image-grounded Qwen3-VL-Embed utility helps bridge this gap without directly optimising these labels, suggesting that multimodal grounding provides complementary signals beyond text consensus. This advantage is amplified by fine-tuning, improving downstream selection performance.

\subsection{Pool Quality Analysis}
\label{sec:pool-quality}

\begin{figure*}[ht]
  \centering
  \includegraphics[width=\linewidth]{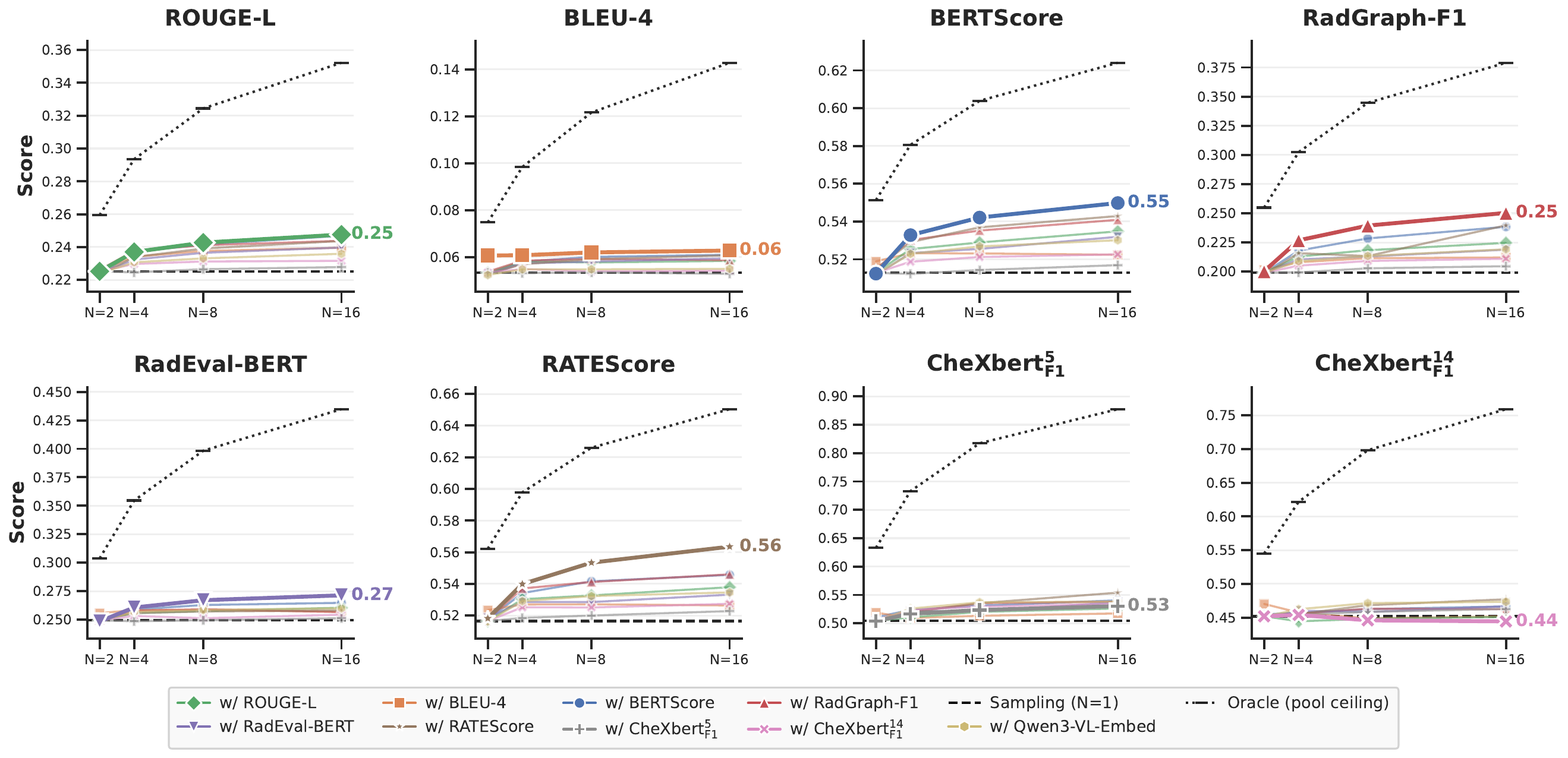}
  \vspace{-25pt}
    \caption{\textbf{Effect of rollout size under different utilities.}
    Each subplot reports one metric as the sampling rollout size varies over $N{\in}\{2,4,8,16\}$ under different consensus utilities.
    Beam-search shows a similar trend in Figure~\ref{fig:metric_trend_beam}.}
  \label{fig:metric_trend_sampling}
  \vspace{-16pt}
\end{figure*}

\paragraph{Pool-Bounded Oracle Ceiling.}
We also report a metric-specific pool-bounded oracle, where for each image and metric, the oracle selects the candidate with the highest reference-based score~\footnote{
The Oracle does not correspond to a single selected report that is optimal across all metrics, but instead reflects the upper bound of the rollout pool under each metric separately.
}.
As shown in Figure~\ref{fig:metric_trend_sampling}, the rollout pool contains reports substantially better than the single output.
This observation echoes prior findings that automated report generation remains far from solved~\citep{zhang2025automated}, but indicates a concrete inference-time opportunity. 
The gap between Sampling and Oracle suggests selection is a critical bottleneck, and that \textsc{CCS} offers a parameter-free inference-time solution. 
Additional results on beam search and decoding temperature are provided in Appx.~\ref{app:other_exp}.

\paragraph{Scaling with Rollout Size.}
Figure~\ref{fig:metric_trend_sampling} also shows that increasing the rollout size improves selection performance, indicating a test-time scaling trend, although the marginal gains taper off. We therefore use $N{=}8$ as a practical trade-off, balancing selection quality with test-time computational cost.

\begin{figure}[ht]
  \centering
  \vspace{-2pt}
  \includegraphics[width=\linewidth]{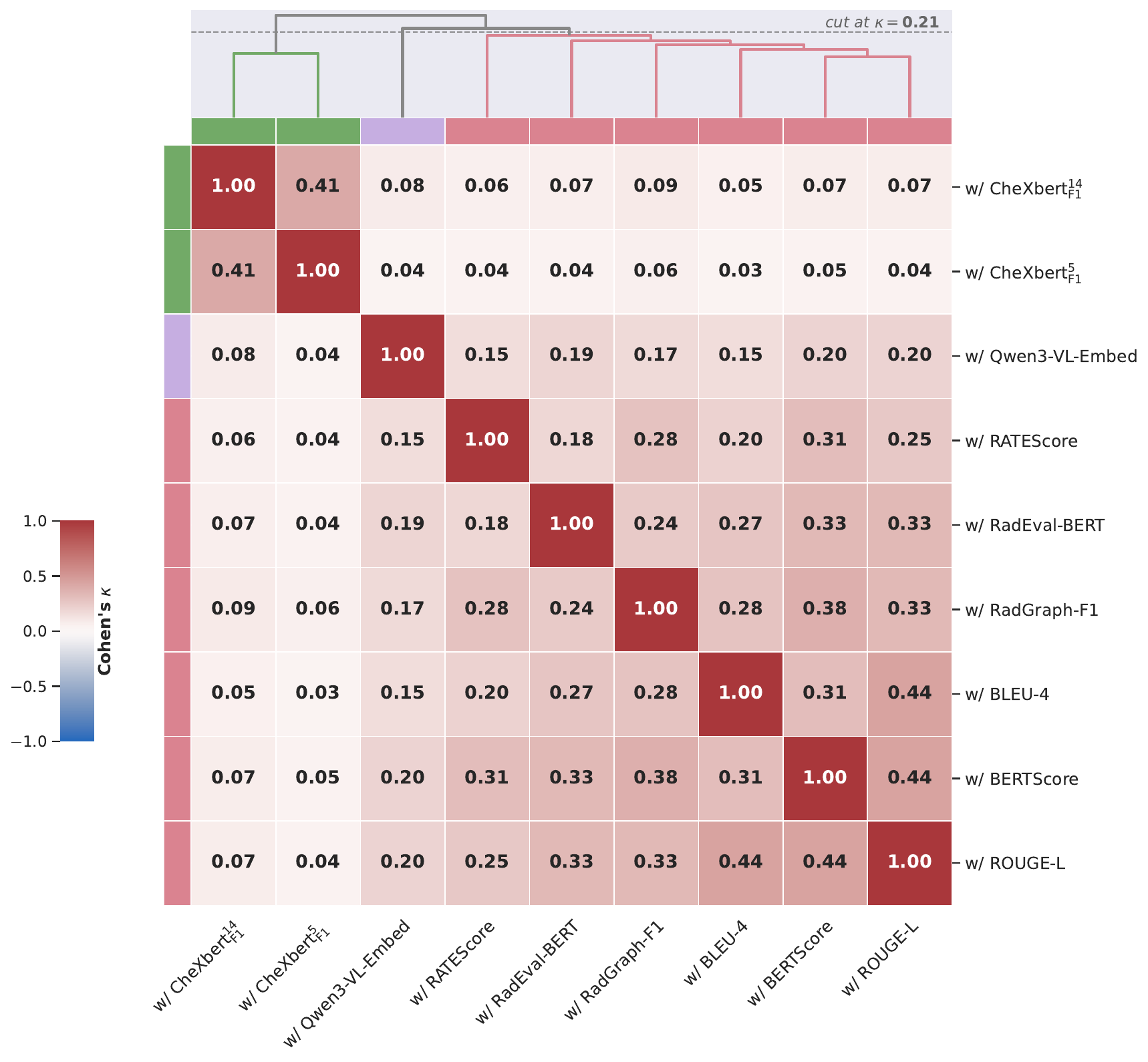}
  \vspace{-25pt}
    \caption{\textbf{Utility decision-space clustermap at $N{=}8$.}
    Pairwise Cohen's $\kappa$ measures agreement between utilities over per-sample candidate choices.
    Hierarchical clustering separates utility groups at $\kappa{=}0.21$.}
  \label{fig:clustermap}
  \vspace{-15pt}
\end{figure}

\subsection{Consensus Geometry Analysis}
\label{sec:geometry}
Consensus utilities make substantially different selection decisions. Clustering pairwise Cohen's $\kappa$~\citep{mchugh2012interrater} over per-sample candidate choices reveals three regimes (Figure~\ref{fig:clustermap}): a semantic cluster covering most text utilities, a clinical-label cluster formed by the two CheXbert variants, and the image-grounded Qwen3-VL-Embed utility as a singleton. The dendrogram cut at $\kappa{=}0.21$, the \emph{slight}--\emph{fair} boundary on the Landis--Koch scale~\citep{landis1977measurement}, separates within-cluster agreement from slight or near-chance cross-cluster agreement. 
Qwen3-VL-Embed induces selections distinct from text- and label-based utilities, consistent with symptom-finding gains not reproduced by either consensus. This clustering reflects utility disagreement, not direct visual grounding.

\input{tables/labels_f1}

\paragraph{Silence Bias of Text Consensus.}

The clinical-label cluster further reflects the ``no finding'' bias discussed above. Because normal or absent findings are more frequent and easier to agree on, label-based consensus may favour conservative candidates with fewer abnormalities. This can improve apparent label agreement while missing clinically important positive findings. By comparison, the image-grounded utility is less affected by this bias: Table~\ref{tab:labels_f1} shows improved per-label F1 across major findings, suggesting better preservation of abnormalities that text-only consensus tends to suppress.

\subsection{Qualitative Analysis}
\label{sec:case-study}

\input{tables/case}

Table~\ref{fig:case} presents a real MIMIC-CXR test case to illustrate how different inference-time strategies affect report quality. Sampling and Greedy represent conventional single-path decoding, whereas \textsc{CCS} introduces consensus-based report selection at inference time without changing the parameters or generation process of the underlying MLLM.

Both single-path baselines exhibit meaningful failure modes. Sampling introduces unsupported statements such as a {\color{gray}prominent mediastinal contour} and {\color{gray}clear lungs}, despite evidence of pulmonary edema and opacity in the reference report. Greedy decoding preserves some major findings but overstates {\color{gray}cardiac enlargement} and incorrectly {\color{gray}localises the catheter tip}. In both cases, clinically relevant observations are either omitted or distorted.

By comparison, \textsc{CCS} produces a more image-grounded and clinically coherent report, preserving cardiac enlargement, pulmonary vascular congestion, edema, and the absence of effusion and pneumothorax, while avoiding the factual errors observed in the baselines. This improvement is reflected in the structured metrics and aligns with the symptom-label analysis in \S\ref{sec:geometry} and Table~\ref{tab:labels_f1}. More broadly, this example supports our observation that radiology report generation remains improvable at inference time, and that clinically stronger reports can be recovered without additional training.

%% file: tables/main_result_part_1.tex
\begin{table*}[ht]
\begin{center}
\small
\renewcommand{\arraystretch}{0.9}
\setlength{\tabcolsep}{2pt}
\resizebox{\textwidth}{!}{%
    \begin{tabular}{l|ccc|c
    c
    c
    c
    c}
        \toprule
        \multirow{3}{*}{\makecell{\textbf{Method}}} & \multicolumn{3}{c|}{\textbf{Lexical Metric}} & \multicolumn{5}{c}{\textbf{Radiology-specific Metric}} \\
        \cmidrule(lr){2-4} \cmidrule(lr){5-9}
        & {ROUGE-L} & {BLEU-4} & {BERTScore} & {RadGraph-F1} & {RaTEScore} & {RadEval-BERT} & {$\text{CheXbert}_{\text{F1}}^{\text{5}}$} & {$\text{CheXbert}_{\text{F1}}^{\text{14}}$} \\
        \midrule
        \rowcolor{gray!15}
        \multicolumn{9}{c}{\textbf{Single Path}} \\
        \midrule
          \text{Sampling} & 0.2252 & 0.0534 & 0.5128 & 0.1989 & 0.5165 & 0.2493 & 0.5041 & 0.4519 \\
          \text{Greedy} & 0.2310 & 0.0538 & 0.5065 & 0.1877 & 0.5192 & 0.2473 & 0.4968 & 0.4109 \\
        \midrule
        \rowcolor{gray!15}
        \multicolumn{9}{c}{\textbf{Rollout (N=8)}} \\
        \midrule
        \text{Random} & 0.2265 & 0.0555 & 0.5150 & 0.2005 & 0.5197 & 0.2521 & 0.5026 & 0.4460 \\
        \text{Perplexity} & 0.2368 & \textbf{0.0694} & \textbf{0.5368} & 0.2125 & 0.5295 & 0.2556 & 0.5148 & 0.4605 \\
        \text{Self-Certainty} & 0.1974 & 0.0328 & 0.4492 & 0.1527 & 0.4664 & 0.2289 & 0.4515 & 0.3990 \\
        \text{ModeX} & \textbf{0.2388} & 0.0595 & 0.5268 & 0.2124 & 0.5291 & 0.2577 & 0.5154 & 0.4496 \\
        \midrule
        \textbf{CCS} & & & & & & & & \\
          \rowcolor{QwenPurple!15}
          \hspace{0.1em} + {Qwen3-VL-Embed} & 0.2331 & 0.0548  & 0.5268 & \textbf{0.2134} & \textbf{0.5323} & \textbf{0.2585} &  \textbf{0.5370} & \textbf{0.4714} \\
          {$p$-value (vs. Sampling)} & 0.0001 & 0.0330 &  0.0001 & 0.0001 & 0.0001 & 0.0006 & 0.0218 & 0.0001  \\
          \hspace{1.75em} {$95\%$ CI of $\Delta$} & {\tiny [+0.0045, +0.0112]} & {\tiny [+0.0003, +0.0055] } & {\tiny [+0.0105, +0.0176] } & { \tiny [+0.0096, +0.0194] }& {\tiny [+0.0119, +0.0199] } & {\tiny [+0.0043, +0.0140] } & {\tiny [+0.0013, +0.0170] } & {\tiny [+0.0070, +0.0151] } \\
        \bottomrule
    \end{tabular}
}
\end{center}
\vspace{-12pt}
\caption{\textbf{Evaluation results on the MIMIC-CXR test split.}
All rollout-based methods select from the same candidate pool with $N=8$, generated with identical MLLM settings, temperature, and random seed.
$p$-values and $95\%$ CIs compare our method against the Sampling baseline.
The best result in each column is shown in \textbf{bold}.}
\label{tab:main_result_1}
\vspace{-10pt}
\end{table*}

%% file: tables/other_MLLMs.tex
\begin{table*}[ht]
\begin{center}
\renewcommand{\arraystretch}{0.8}
\small
\setlength{\tabcolsep}{2pt}
\resizebox{\textwidth}{!}{
    \begin{tabular}{l|ccc|c
    c
    c
    c
    c}
        \toprule
        \multirow{3}{*}{\makecell{\textbf{Method}}} & \multicolumn{3}{c|}{\textbf{Lexical Metric}} & \multicolumn{5}{c}{\textbf{Radiology-specific Metric}} \\
        \cmidrule(lr){2-4} \cmidrule(lr){5-9}
        & {ROUGE-L} & {BLEU-4} & {BERTScore} & {RadGraph-F1} & {RaTEScore} & {RadEval-BERT} & {$\text{CheXbert}_{\text{F1}}^{\text{5}}$} & {$\text{CheXbert}_{\text{F1}}^{\text{14}}$} \\
        \midrule
        \rowcolor{gray!15}
        \multicolumn{9}{c}{\textbf{MIMIC-CXR}} \\
        \midrule
          \textbf{LLaVA-Med} & 0.1479 & 0.0090 & 0.3758 & 0.0723 & 0.4292 & 0.1768 & 0.2492 & 0.2282 \\
           \rowcolor{QwenPurple!15} \hspace{0.1em} \textbf{\textsc{\(+\) CCS}} & \underline{0.1514}\up{} & \textbf{0.0098}\up{} & \underline{0.3845}\up{} & \textbf{0.0766}\up{} & 0.4341\up{} & 0.1773\up{} & \underline{0.2546}\up{} & \textbf{0.2401}\up{} \\
        \midrule
          \textbf{LLaVA-Rad} & 0.2396 & 0.0700 & 0.5271 & 0.2128 & 0.5342 & 0.2903 & 0.5706 & 0.5406 \\
           \rowcolor{QwenPurple!15} \hspace{0.1em} \textbf{\textsc{\(+\) CCS}} & \underline{0.2484}\up{} & \textbf{0.0767}\up{} & 0.5319\up{} & \underline{0.2216}\up{} & 0.5409\up{} & \underline{0.2977}\up{} & \textbf{0.6014}\up{} & \underline{0.5619}\up{} \\
        \midrule
          \textbf{Libra} & 0.2091 & 0.0462 & 0.5024 & 0.1918 & 0.5248 & 0.2597 & 0.5785 & 0.5146 \\
           \rowcolor{QwenPurple!15} \hspace{0.1em} \textbf{\textsc{\(+\) CCS}} & 0.2106\up{} & 0.0430\down{} & 0.5018\down{} & \underline{0.1955}\up{} & 0.5258\up{} & 0.2635\up{} & \underline{0.5988}\up{} & \underline{0.5351}\up{} \\
        \midrule
        \rowcolor{gray!15}
        \multicolumn{9}{c}{\textbf{IU-Xray}} \\
        \midrule
          \textbf{LLaVA-Med} & 0.1218 & 0.0038 & 0.3399 & 0.0696 & 0.4212 & 0.2005 & 0.0639 & 0.0588 \\
           \rowcolor{QwenPurple!15} \hspace{0.1em} \textbf{\textsc{\(+\) CCS}} & \underline{0.1251}\up{} & \underline{0.0039}\up{} & \underline{0.3471}\up{} & 0.0706\up{} & 0.4250\up{} & 0.2017\up{} & \textbf{0.0701}\up{} & 0.0591\up{} \\
        \midrule
          \textbf{LLaVA-Rad} & 0.2243 & 0.0381 & 0.4785 & 0.2128 & 0.5563 & 0.2142 & 0.4197 & 0.4732 \\
           \rowcolor{QwenPurple!15} \hspace{0.1em} \textbf{\textsc{\(+\) CCS}} & 0.2243 & \textbf{0.0398}\up{} & 0.4743\down{} & 0.2129\up{} & 0.5608\up{} & 0.2150\up{} & 0.4268\up{} & 0.4772\up{} \\
        \midrule
          \textbf{Libra} & 0.2362 & 0.0304 & 0.4763 & 0.2650 & 0.5367 & 0.2431 & 0.4097 & 0.4595 \\
           \rowcolor{QwenPurple!15} \hspace{0.1em} \textbf{\textsc{\(+\) CCS}} & 0.2386\up{} & 0.0279\down{} & 0.4771\up{} & 0.2694\up{} & 0.5374\up{} & 0.2462\up{} & \textbf{0.4578}\up{} & \textbf{0.4822}\up{} \\
        \midrule
        \rowcolor{gray!15}
        \multicolumn{9}{c}{\textbf{CheXpert Plus}} \\
        \midrule
          \textbf{LLaVA-Med} & 0.1417 & 0.0091 & 0.3622 & 0.0822 & 0.4204 & 0.1780 & 0.3201 & 0.2865 \\
           \rowcolor{QwenPurple!15} \hspace{0.1em} \textbf{\textsc{\(+\) CCS}} & 0.1404\down{} & \textbf{0.0103}\up{} & 0.3451\down{} & \textbf{0.0862}\up{} & 0.4281\up{} & 0.1812\up{} & 0.3231\up{} & \underline{0.2977}\up{} \\
        \midrule
          \textbf{LLaVA-Rad} & 0.1827 & 0.0197 & 0.4355 & 0.1557 & 0.4725 & 0.2317 & 0.4904 & 0.5007 \\
           \rowcolor{QwenPurple!15} \hspace{0.1em} \textbf{\textsc{\(+\) CCS}} & \underline{0.1886}\up{} & \textbf{0.0297}\up{} & 0.4365\up{} & \underline{0.1588}\up{} & 0.4753\up{} & \textbf{0.2550}\up{} & \textbf{0.5456}\up{} & \textbf{0.5474}\up{} \\
        \midrule
          \textbf{Libra} & 0.1933 & 0.0248 & 0.4767 & 0.1877 & 0.4980 & 0.2660 & 0.5052 & 0.5498 \\
           \rowcolor{QwenPurple!15} \hspace{0.1em} \textbf{\textsc{\(+\) CCS}} & 0.1925\down{} & 0.0213\down{} & \underline{0.4880}\up{} & \textbf{0.2261}\up{} & \underline{0.5165}\up{} & \textbf{0.2772}\up{} & \textbf{0.5728}\up{} & 0.5586\up{} \\
        \bottomrule
    \end{tabular}
}
\end{center}
\vspace{-12pt}
\caption{\textbf{Evaluation results across radiology MLLM backbones and datasets.}
\textsc{CCS} uses Qwen3-VL-Embed as the clinical consensus utility.
All rollout pools are generated with sampling temperature $\tau=0.5$ and pool size $N=8$.
``\up{}\;\;'' and ``\down{}\;\;'' indicate changes relative to the corresponding sampling baseline.
Within each ``$+$ \textsc{CCS}'' row, metrics are marked by the empirical distribution of relative changes $\delta=(\textsc{CCS}-\text{baseline})/\text{baseline}$: \textbf{bold} indicates upper-quartile gains ($\delta \ge {+}4.17\%$), while \underline{underline} indicates median-to-upper-quartile gains ($\delta \ge {+}1.88\%$).
}
\label{tab:other_mllms}
\vspace{-15pt}
\end{table*}

%% file: tables/main_result_part_2.tex
\begin{table*}[ht]
\begin{center}
\small
\renewcommand{\arraystretch}{0.9}
\setlength{\tabcolsep}{2pt}
\resizebox{\textwidth}{!}{%
    \begin{tabular}{l|ccc|c
    c
    c
    c
    c}
        \toprule
        \multirow{3}{*}{\makecell{\textbf{Method}}} & \multicolumn{3}{c|}{\textbf{Lexical Metric}} & \multicolumn{5}{c}{\textbf{Radiology-specific Metric}} \\
        \cmidrule(lr){2-4} \cmidrule(lr){5-9}
        & {ROUGE-L} & {BLEU-4} & {BERTScore} & {RadGraph-F1} & {RaTEScore} & {RadEval-BERT} & {$\text{CheXbert}_{\text{F1}}^{\text{5}}$} & {$\text{CheXbert}_{\text{F1}}^{\text{14}}$} \\
        \midrule
        \rowcolor{gray!15}
        \multicolumn{9}{c}{\textbf{Textual Utility}} \\
        \midrule
          \rowcolor{LightCyan!15}
          \hspace{0.1em} + {ROUGE-L} & {\bfseries 0.2427} & 0.0577 & 0.5289 & 0.2183 & 0.5327 & 0.2575 & 0.5202 & 0.4481 \\
          \rowcolor{LightCyan!15}
          \hspace{0.1em} + {BLEU-4} & 0.2376 & {\bfseries 0.0620} & 0.5231 & 0.2115 & 0.5271 & 0.2584 & 0.5133 & 0.4488 \\
          \rowcolor{LightCyan!15}
          \hspace{0.1em} + BERTScore & 0.2415 & 0.0601 & {\bfseries 0.5421} & 0.2284 & 0.5416 & 0.2628 & 0.5312 & 0.4625 \\
          \rowcolor{LightCyan!15}
          \hspace{0.1em} + {RadGraph-F1} & 0.2411 & 0.0592 & 0.5352 & {\bfseries 0.2394} & 0.5412 & 0.2591 & 0.5357 & 0.4631 \\
          \rowcolor{LightCyan!15}
          \hspace{0.1em} + {RATEScore} & 0.2391 & 0.0591 & 0.5369 & 0.2133 & {\bfseries 0.5534} & 0.2571 & 0.5355 & 0.4683 \\
          \rowcolor{LightCyan!15}
          \hspace{0.1em} + {RadEval-BERT} & 0.2365 & 0.0581 & 0.5255 & 0.2129 & 0.5285 & {\bfseries 0.2670} & 0.5211 & 0.4583 \\
          \rowcolor{LightCyan!15}
          \hspace{0.1em} + {$\text{CheXbert}_{\text{F1}}^{\text{5}}$} & 0.2265 & 0.0535 & 0.5143 & 0.2028 & 0.5200 & 0.2494 & 0.5234 & 0.4584 \\
          \rowcolor{LightCyan!15}
          \hspace{0.1em} + {$\text{CheXbert}_{\text{F1}}^{\text{14}}$} & 0.2312 & 0.0540 & 0.5212 & 0.2091 & 0.5251 & 0.2512 & 0.5295 & 0.4459 \\
         \midrule
        \rowcolor{gray!15}
        \multicolumn{9}{c}{\textbf{Image-Grounded Utility}} \\
        \midrule
          \rowcolor{QwenPurple!15}
          \hspace{0.1em} + {Qwen3-VL-Embed} & 0.2331 & 0.0548  & 0.5268 & 0.2134 & 0.5323 & 0.2585 & {\bfseries 0.5370} & {\bfseries 0.4714} \\
          \hspace{0.6em} $\boldsymbol{\hookrightarrow}$ {w/o Fine-tuning} & 0.2375 & 0.0601  & 0.5356 & 0.2113 & 0.5295 & 0.2536 & 0.5332 & 0.4700 \\
        \bottomrule
    \end{tabular}
}
\end{center}
\vspace{-12pt}
\caption{\textbf{Comparison of \textsc{CCS} with different consensus utilities.}
All utilities select from the same rollout pool, isolating the effect of the consensus scoring function.
The `w/o fine-tuning' variant is the original Qwen3-VL-Embed checkpoint before radiology-specific adaptation for RRG.
The best result in each column is shown in \textbf{bold}.
}
\label{tab:main_result_2}
\vspace{-12pt}
\end{table*}

%% file: tables/labels_f1.tex
\begin{table}[h]
    \centering
    \small
    \vspace{-5pt}
    \renewcommand{\arraystretch}{0.9}
    \setlength{\tabcolsep}{2pt}
    \resizebox{0.9\linewidth}{!}{
    \begin{tabular}{l|ccc}
        \toprule
        \textbf{Metric} & Sampling & \textit{w/} \textsc{CCS} & $\Delta$ \\
        \midrule

        \rowcolor{gray!15}
        \multicolumn{4}{c}{\textbf{CheXbert-F1} \textit{(5-class)}} \\
        \midrule
         Atelectasis & 0.4215 & 0.4544 & +0.0329 \\
          Cardiomegaly & 0.5968 & 0.6204 & +0.0236 \\
          Consolidation & 0.1242 & 0.1514 & +0.0272 \\
          Edema & 0.4390 & 0.4803 & +0.0413 \\
          Pleural Effusion & 0.6122 & 0.6510 & +0.0388 \\
        \bottomrule
    \end{tabular}
    }
    \vspace{-5pt}
    \caption{\textbf{CheXbert 5-class F1 comparison by symptom label.}
    Sampling vs.\ \textsc{CCS} with Qwen3-VL-Embed utility on MIMIC-CXR.
    $\Delta$ denotes the absolute F1 gain.}
    \label{tab:labels_f1}
    \vspace{-13pt}
\end{table}

%% file: tables/case.tex
\providecommand{\blueul}[1]{{\color{blue}\underline{\textcolor{black}{#1}}}}

\begin{table*}[ht]
\small
\centering
\begin{tabular}{@{}p{0.07\linewidth}p{0.9\linewidth}@{}}
\toprule
\textbf{Ground Truth} & A portable AP upright view of the chest was obtained. Again seen is a \blueul{right-sided dialysis catheter} terminating in the right atrium. \textcolor{blue}{Heart is mildly enlarged}. \textcolor{blue}{Pulmonary vasculature is mildly engorged}. A \blueul{rounded opacity at the right base}, present sicne \_\_\_, may represent \textcolor{blue}{asymmetric pulmonary edema}, but other processes such as \blueul{pulmonary abscess} cannot be excluded. No large \textcolor{blue}{effusion}, or \blueul{pneumothorax}.
 \\
\midrule
\textbf{Sampling} & AP portable view of the chest demonstrates interval placement of a \blueul{right internal jugular central venous catheter}, tip at cavoatrial junction. No \blueul{pneumothorax}. No \textcolor{blue}{pleural effusion}. \textcolor{blue}{Stable mild cardiomegaly}. \textcolor{red}{Stably prominent mediastinal contour}. \textcolor{red}{The lungs are clear}. \blueul{Surgical clips} project over right upper abdomen.
\newline
{\color{gray}\textit{\hspace{12em} RadGraph-F1 = 0.1304 \hspace{3em} $\text{CheXbert}_{\text{F1}}^{\text{5}}$ = 0.5000 \hspace{3em} 
{$\text{CheXbert}_{\text{F1}}^{\text{14}}$} = 0.6667}} \\
\midrule
\textbf{Greedy} & In comparison with the study of \_\_\_, there is little overall change. Again there is \textcolor{red}{substantial} \textcolor{blue}{enlargement of the cardiac silhouette} with some \textcolor{blue}{elevation of pulmonary venous pressure}. The \blueul{right IJ catheter} tip is in the \textcolor{red}{mid portion of the SVC}. No evidence of acute focal \blueul{pneumonia}.
\newline
{\color{gray}\textit{\hspace{12em} RadGraph-F1 = 0.0513 \hspace{3em}
$\text{CheXbert}_{\text{F1}}^{\text{5}}$ = 0.0000 \hspace{3em}
{$\text{CheXbert}_{\text{F1}}^{\text{14}}$} = 0.3333}}\\
\midrule
\textbf{\textsc{CCS}} & \mbox{}\blueul{Right-sided dialysis catheter} is unchanged in position. \textcolor{blue}{The heart is moderately enlarged}, and there is \textcolor{blue}{mild pulmonary vascular congestion} and \textcolor{blue}{pulmonarily edema}. There is no focal \textcolor{blue}{consolidation} or \textcolor{blue}{pleural effusion}. No \blueul{pneumothorax}. \blueul{Surgical clips} are noted in the right upper quadrant.
\newline
{\color{gray}\textit{\hspace{12em} RadGraph-F1 = 0.2857 \hspace{3em} $\text{CheXbert}_{\text{F1}}^{\text{5}}$ = 1.0000 \hspace{3em} {$\text{CheXbert}_{\text{F1}}^{\text{14}}$} = 1.0000}} \\
\bottomrule
\end{tabular}
\vspace{-8pt}
\caption{\textbf{Qualitative case study.}
Comparison of Sampling, Greedy, and \textsc{CCS} with Qwen3-VL-Embed utility on a MIMIC-CXR test sample (\texttt{study\_id}: \texttt{54124205}, \texttt{subject\_id}: \texttt{17340686}).
\textcolor{blue}{Blue} text marks CheXbert-5 findings, \blueul{blue underlining} marks additional CheXbert-14 findings, and \textcolor{red}{red} text marks factual errors against the ground truth.}
\vspace{-15pt}
\label{fig:case}
\end{table*}

%% file: sections/conclusion.tex
\section{Conclusion}
\label{sec:conclusion}

We introduce \textbf{\textsc{CCS}}, a reference-free, decoder-agnostic inference-time selection framework that reframes radiology report generation as candidate selection over a rollout pool. Given candidates from a fixed MLLM, \textsc{CCS} selects the report with the highest clinical consensus. Across three datasets and multiple backbones, \textsc{CCS} consistently improves clinical report quality over single-path decoding without retraining. These results show that radiology MLLMs can often generate better reports than those they initially commit to, and that image-grounded utility can help recover them.


%% file: sections/limitations.tex
\section*{Limitations}
\label{sec:limitations}

Several limitations remain. First, our experiments are conducted on standard radiology benchmark datasets with curated image--report pairs. Although these datasets are widely adopted for evaluating RRG systems, they may not fully capture the diversity and noise encountered in real-world clinical workflows, including variations in acquisition protocols and reporting styles. Second, our evaluation relies on automatic clinical metrics and does not include assessment by licensed radiologists. While expert evaluation is particularly important for rigorous validation in medical domains, conducting large-scale clinical studies remains outside the scope of this work. Third, we do not include LLM-as-a-judge evaluation or explore larger multimodal embedding backbones for consensus estimation. Although our results suggest that image-grounded utilities provide useful selection signals, additional validation strategies and stronger embedding models may offer complementary evidence and further improve candidate selection.

%% file: sections/ethical.tex
\section*{Ethical Considerations}

This work uses only publicly available, de-identified radiology datasets and follows the corresponding dataset usage policies and licences. No private patient information is used. The IDs reported in the caption of Table~\ref{fig:case} are official timestamp-based identifiers provided by the dataset and do not contain patient-identifiable information. Our method is intended for research on assistive radiology AI, rather than autonomous clinical decision-making. Any practical use of radiology report generation systems should involve licensed clinicians, appropriate validation, and careful monitoring to avoid over-reliance on automated outputs.

%% file: sections/appendix.tex
\section{Research Objectives}
\label{app:objective}
 

\subsection{Research Aims}
\label{app:objective_aims}
This work introduces \textbf{\textsc{CCS}} (Clinical Consensus Selection), a reference-free and decoder-agnostic Best-of-\textit{N} framework for radiology report generation (RRG). The primary objective is to improve the clinical quality of generated reports \emph{at inference time}, by selecting a more clinically reliable report from a pool of candidates sampled from a fixed radiology MLLM, without modifying model parameters, retraining the generator, or relying on external corpora.

It is equally important to clarify what this work does not aim to address. We do not propose a new generation architecture or training algorithm, nor do we seek to improve the generator itself; our focus is on the \emph{selection} stage applied to already-generated candidates. Consequently, we do not compare against methods that require architectural modifications, additional supervised training, or retrieval-based augmentation from external knowledge bases. \textsc{CCS} is instead complementary to such approaches: any generator, including one improved through these means, can serve as the backbone from which candidates are drawn.

\subsection{Research Scope}
\label{app:objective_scope}
This study focuses on report generation for chest X-rays, the most widely used imaging modality in clinical practice. All experiments use frontal-view radiographs only, namely anterior--posterior (AP) and posterior--anterior (PA) projections, and target the generation of the \textit{Findings} section. We evaluate on three public datasets---MIMIC-CXR, IU-Xray, and CheXpert Plus---where models are trained only on MIMIC-CXR and evaluated on the other two to assess cross-dataset generalisation. To examine whether the framework generalises across generators, we apply \textsc{CCS} to several pre-trained radiology MLLMs, including LLaVA-Med, LLaVA-Rad, and Libra, in addition to our baseline MLLM. The image-grounded utility is obtained by adapting a multimodal embedding model (Qwen3-VL-Embed) to CXR--report representation learning on the same training data.

Several directions are intentionally left outside our scope. We do not address other imaging modalities such as computed tomography (CT), magnetic resonance imaging (MRI), or ultrasound, nor do we incorporate auxiliary signals from clinical notes, laboratory values, or electronic health records. We also do not modify the generation process of the underlying MLLM or apply post-hoc report rewriting; \textsc{CCS} operates entirely as an inference-time selection step over candidates produced by an unmodified generator, which keeps it compatible with a wide range of pre-trained models at low deployment cost.

\section{Dataset and Metrics}
\subsection{Dataset Description}
\label{app:datasets}
\paragraph{MIMIC-CXR}~\citep{johnson2019mimic}\quad
MIMIC-CXR is a large-scale publicly available chest radiography dataset, comprising 377,110 chest radiographs from 227,835 imaging studies, each paired with a free-text radiology report. We use the JPEG images from the MIMIC-CXR-JPG release~\citep{johnson2019mimiccxrjpglargepubliclyavailable}, which are derived from the original DICOM files. To ensure consistency across datasets, we retain only frontal-view images, including anterior-posterior (AP) and posterior-anterior (PA) views.

Each report is preprocessed to extract clinically relevant sections, including \textit{Findings}, \textit{Indication}, \textit{Technique}, \textit{Comparison}, and \textit{History}. This is performed using pattern-matching heuristics adapted from the official preprocessing scripts~\citep{johnson2018mimic}. For training, we use only the MIMIC-CXR training split: both the backbone MLLM and Qwen3-VL-Embed are trained on 162,955 training records, with 1,286 records used for validation. No IU-Xray or CheXpert Plus samples are used for training, allowing evaluation on these datasets to reflect cross-dataset generalisation. For evaluation, we report results on the official test split, consisting of 2,461 studies with frontal-view images and non-empty \textit{Findings} sections.

\paragraph{IU-Xray}~\citep{demner2015preparing}\quad
IU-Xray is a publicly available chest X-ray dataset for medical image analysis and radiology report generation, containing 7,470 chest X-ray images and 3,955 corresponding diagnostic reports. All images are converted to PNG format. For evaluation, we select 3,307 frontal-view cases with non-empty \textit{Findings} sections.

\paragraph{CheXpert Plus}~\citep{chambon2024chexpert}\quad
CheXpert Plus is a large-scale chest radiography dataset comprising 223,462 image--report pairs from 187,711 studies across 64,725 patients. As the official test split is not publicly available, we evaluate on the public validation set. After filtering for frontal-view images with non-empty \textit{Findings} sections, the resulting evaluation set contains 62 samples.

\subsection{Evaluation Metrics}
\label{app:metrics}

\paragraph{Lexical Metrics}

We use standard natural language generation metrics to evaluate textual similarity between generated and reference reports. ROUGE-L~\citep{lin-2004-rouge} measures the longest common subsequence, BLEU-4~\citep{10.3115/1073083.1073135} computes n-gram ($n=4$) precision with a brevity penalty, and BERTScore~\citep{zhang2020bertscoreevaluatingtextgeneration} estimates semantic similarity using contextual embeddings from BERT~\citep{devlin2019bertpretrainingdeepbidirectional}. All metrics are computed with their default configurations.

\paragraph{Radiology-specific Metrics}

We adopt several radiology-specific metrics to assess the clinical correctness of generated reports.\footnote{For fairness, reproducibility, and consistency with prior work, all lexical and radiology-specific metrics are computed using the RadEval toolkit~\citep{xu-etal-2025-radeval}, version \texttt{0.0.6rc2}, with default configurations.} RadGraph-F1~\citep{delbrouck-etal-2022-improving} represents reports as structured graphs of clinical entities, such as anatomical sites and observations, and their relations. RaTEScore~\citep{zhao2024ratescoremetricradiologyreport} evaluates critical diagnostic concepts and anatomical details, while accounting for medical synonyms and negation cues. RadEval-BERT~\citep{xu-etal-2025-radeval} uses a radiology-adapted ModernBERT model~\citep{warner2024smarterbetterfasterlonger} to measure semantic similarity between generated and reference reports. CheXbert-F1~\citep{smit2020chexbertcombiningautomaticlabelers} applies an automatic labeler to extract ``present'', ``absent'', and ``uncertain'' labels for 14 clinical conditions~\citep{irvin2019chexpertlargechestradiograph}. We report the weighted F1 score for both the full 14-class setting and the 5-class setting. The 5-class setting focuses on five common pathologies: \textit{Atelectasis}, \textit{Cardiomegaly}, \textit{Consolidation}, \textit{Edema}, and \textit{Pleural Effusion}.

\section{Experimental Details}
\label{app:experimental_details}

This section provides additional experimental details, including the training configurations of the baseline MLLMs and the multimodal embedding model, the prompt templates used in our experiments, and brief descriptions of the three pre-trained radiology MLLMs evaluated in this work.

All model training and experiments are conducted on a single NVIDIA A6000 GPU with 48GB memory. Although \textsc{CCS} requires multiple rollout generations at inference time, it introduces only moderate deployment overhead, as modern Transformer libraries support efficient batched inference. In our implementation, compared with single-candidate decoding, batched rollout generation takes approximately $1.4\times$, $2.0\times$, and $3.0\times$ runtime for $N=4$, $N=8$, and $N=16$, respectively. The actual runtime may vary with the hardware configuration, particularly the available GPU floating-point throughput.

\subsection{Training Details}
\label{app:train_details}

This section provides the training details for the two trainable components used in our experiments: the baseline MLLM for report generation and the Qwen3-VL-Embed-2B model for CXR--report representation learning. Both models are trained using the same training split described in Appx.~\ref{app:datasets}, but they optimise different objectives and therefore use different training configurations.

Specifically, the baseline MLLM is trained for conditional report generation using the standard autoregressive language-modelling objective:
\begin{equation}
\mathcal{L}_{\mathrm{gen}}
=
-\frac{1}{T}
\sum_{t=1}^{T}
\log p_{\theta}(y_t \mid y_{<t}, x, q),
\end{equation}
where $x$ denotes the input CXR image, $q$ denotes the instruction, and $y=\{y_t\}_{t=1}^{T}$ denotes the report.

In contrast, Qwen3-VL-Embed-2B is adapted for CXR--report representation learning using an instruction-conditioned InfoNCE objective. Each training instance is formulated as a query--target pair $(\mathbf{q}_i,\mathbf{t}_i^+)$, where $\mathbf{q}_i$ denotes the instruction-prefixed query and $\mathbf{t}_i^+$ denotes its matched report. Given a mini-batch of $B$ query--target pairs, we first define the temperature-scaled similarity score as
\begin{equation}
s_{ij}
=
\cos(\mathbf{h}_{q_i}, \mathbf{h}_{t_j}) / \tau,
\end{equation}
and optimise the InfoNCE objective:
\begin{equation}
\mathcal{L}_{\mathrm{InfoNCE}}
=
-\frac{1}{B}
\sum_{i=1}^{B}
\log
\frac{
\exp(s_{ii})
}{
\sum_{j=1}^{B}
\exp(s_{ij})
},
\end{equation}
where $\mathbf{h}_{q_i}=f_{\theta}(\mathbf{q}_i)$ and $\mathbf{h}_{t_j}=f_{\theta}(\mathbf{t}_j)$ are the query and target embeddings encoded by Qwen3-VL-Embed-2B, $\tau$ is the contrastive temperature, $s_{ii}$ corresponds to the matched query--target pair, and $s_{ij}$ with $j\neq i$ corresponds to in-batch negatives. This contrastive adaptation enables the embedding model to provide an image-grounded utility score for candidate report selection.

Detailed hyperparameters for the two models are summarised in Tables~\ref{table:hyperparam_mllm} and~\ref{table:hyperparam_qwen_embed}, respectively.

\begin{table}[h]
    \centering
    \small
    \resizebox{\linewidth}{!}
    {
    \begin{tabular}{l|cc}
        \toprule
        \textbf{Configuration} & \textbf{Stage I} & \textbf{Stage II} \\
        \midrule
        Base Model & \multicolumn{2}{c}{LLaVA-v1.5-7b} \\
        Training Objective & CXR--text alignment & RRG instruction tuning \\
        Trainable Module & Projector (2-layer MLP) & LLM (LoRA adapters) \\
        \midrule
        Training Epoch & $1$ & $3$ \\
        \midrule
        Learning Rate & \multicolumn{2}{c}{$1 \times 10^{-5}$} \\
        Optimizer & \multicolumn{2}{c}{AdamW} \\
        LR Scheduler & \multicolumn{2}{c}{Cosine} \\
        Warmup Ratio & \multicolumn{2}{c}{$0.03$} \\
        LoRA Config & -- & $r=128, \alpha=256$ \\
        Batch Size & \multicolumn{2}{c}{$16$} \\
        Precision & \multicolumn{2}{c}{BF16} \\
        \bottomrule
    \end{tabular}
    }
    \caption{Detailed hyperparameters for training the baseline MLLM in two stages. Stage I fully fine-tunes the projector for CXR--text alignment with the visual encoder and LLM frozen, while Stage II applies LoRA to fine-tune the LLM for RRG.}
    \label{table:hyperparam_mllm}
\end{table}

\begin{table}[h]
    \centering
    \small
    \resizebox{\linewidth}{!}
    {
    \begin{tabular}{l|c}
        \toprule
        \textbf{Configuration} & \textbf{Single Stage} \\
        \midrule
        Base Model & Qwen3-VL-Embed-2B \\
        Training Objective & CXR--report representation learning \\
        Trainable Module & LoRA adapters \\
        \midrule
        Training Epoch & $1$ \\
        \midrule
        Learning Rate & $1 \times 10^{-4}$ \\
        Optimizer & AdamW \\
        LR Scheduler & Cosine \\
        Warmup Ratio & $0.01$ \\
        LoRA Config & $r=8, \alpha=32$ \\
        Batch Size & Dynamic \\
        Precision & BF16 \\
        Contrastive Temperature & $\tau=0.01$ \\
        False-negative Margin & $\delta=0.1$ \\
        \bottomrule
    \end{tabular}
    }
    \caption{Detailed hyperparameters for adapting Qwen3-VL-Embed-2B for CXR--report representation learning. LoRA adapters are fine-tuned with a contrastive objective, using temperature $\tau$ and false-negative margin $\delta$ for embedding optimisation.}
    \label{table:hyperparam_qwen_embed}
\end{table}

\subsection{Prompt Details}
\label{app:prompt}

We provide the prompt templates used for MLLM-based report generation and Qwen3-VL-Embed representation encoding in Table~\ref{table:prompt_template}. Part~(A) defines the input format for generating candidate findings reports from chest X-ray images and available clinical context, while Part~(B) defines the query and document formats used by Qwen3-VL-Embed for CXR--report representation learning and inference. These templates are used consistently during training and inference.

\begin{table*}[t]
\centering
\small
\resizebox{\textwidth}{!}{
\begin{tabular}{l|l}
    \toprule
    \textbf{Role} & \textbf{Prompt} \\
    \midrule
    \rowcolor{gray!8}
    \multicolumn{2}{c}{\textbf{(A) Multimodal Large Language Models}} \\
    \midrule
    \multirow{4}{*}{\textsc{System}} & \texttt{<|system|>} \\
    & \multicolumn{1}{p{13cm}}{A chat between a curious human and an artificial intelligence assistant. The assistant gives helpful, detailed, and polite answers to the human's questions.} \\
    & \texttt{<|end|>} \\
    \midrule
    \multirow{7}{*}{\textsc{User}} 
    & \texttt{<|user|>} \\
    & {\color{blue}\texttt{<chest X-ray image>}} \\
    & {\color{gray}\textit{Indication}: \ldots \ \ (\textit{if available})} \\
    & {\color{gray}\textit{Technique}: \ldots \ \ (\textit{if available})} \\
    & {\color{gray}\textit{Comparison}: \ldots \ \ (\textit{if available})} \\
    & \multicolumn{1}{p{13cm}}{Provide a detailed description of the findings in the radiology image.} \\
    & \texttt{<|end|>} \\
    \midrule
    \multirow{4}{*}{\textsc{Assistant}} 
    & \texttt{<|assistant|>} \\
    & {\color{gray}(\textit{Findings section})} \ldots  \\
    & {\color{gray}(\textit{e.g., the target})} \\
    & \texttt{<|end|>} \\
    \midrule
    \rowcolor{gray!8}
    \multicolumn{2}{c}{\textbf{(B) Qwen3-VL-Embed}} \\
    \midrule
    \multirow{3}{*}{\textsc{System}} & \texttt{<|system|>} \\
    & \multicolumn{1}{p{13cm}}{Provide a detailed description of the findings in the radiology image.} \\
    & \texttt{<|end|>} \\
    \midrule
    \multirow{6}{*}{\makecell[l]{\textsc{User}\\(Query)}} 
    & \texttt{<|user|>} \\
    & {\color{blue}\texttt{<chest X-ray image>}} \\
    & {\color{gray}\textit{Indication}: \ldots \ \ (\textit{if available})} \\
    & {\color{gray}\textit{Technique}: \ldots \ \ (\textit{if available})} \\
    & {\color{gray}\textit{Comparison}: \ldots \ \ (\textit{if available})} \\
    & \texttt{<|end|>} \\
    \midrule
    \multirow{5}{*}{\makecell[l]{\textsc{User}\\(Document)}} 
    & \texttt{<|user|>} \\
    & \multicolumn{1}{p{13cm}}{Represent the user's input. {\color{gray}\textit{(default instruction) }}}\\
    & {\color{gray}(\textit{Findings section})} \ldots  \\
    & {\color{gray}(\textit{e.g., the target})} \\
    & \texttt{<|end|>} \\
    \bottomrule
\end{tabular}
}
\caption{\textbf{Prompt templates used in this work.} 
The templates include both the report-generation prompt for MLLM rollout and the query/document prompts for Qwen3-VL-Embed representation learning and inference. 
The same templates are used consistently during training and inference unless otherwise specified.}
\label{table:prompt_template}
\end{table*}

\subsection{Pre-trained Radiology Models}
\label{app:mllms}

\paragraph{LLaVA-Med}~\citep{li2023llavamed}\quad
LLaVA-Med is a biomedical extension of LLaVA~\citep{liu2023visualinstructiontuning}, developed to support multimodal instruction following in biomedical domains. It is trained using synthetic instruction-following data derived from PMC-15M~\citep{zhang2025biomedclipmultimodalbiomedicalfoundation} image--text pairs, where GPT-4~\citep{openai2024gpt4technicalreport} is used to generate instructions without manual annotation. The training procedure consists of biomedical vision--language alignment followed by instruction tuning for open-ended biomedical dialogue. In our experiments, we use LLaVA-Med v1.5, which is built with Mistral-7B~\citep{jiang2023mistral7b} as the backbone and a jointly trained CLIP-based visual encoder~\citep{radford2021learningtransferablevisualmodels}. This model provides a general biomedical MLLM baseline for evaluating report generation from chest X-ray images.

\paragraph{LLaVA-Rad}~\citep{Zambrano_Chaves_2025}\quad
LLaVA-Rad is a radiology-oriented instruction-tuned MLLM for chest X-ray report generation. It follows the LLaVA~\citep{liu2023visualinstructiontuning} architecture and uses LoRA~\citep{hu2021loralowrankadaptationlarge} for parameter-efficient adaptation. The model is trained on MIMIC-CXR, using radiology reports that are further structured with GPT-4~\citep{openai2024gpt4technicalreport} to improve consistency and label clarity. For image encoding, LLaVA-Rad employs BiomedCLIP~\citep{zhang2025biomedclipmultimodalbiomedicalfoundation}, a biomedical vision--language encoder pretrained on large-scale biomedical image--text pairs. This design makes LLaVA-Rad a domain-specialised baseline for RRG.

\paragraph{Libra}~\citep{zhang2025libraleveragingtemporalimages}\quad
Libra is a multimodal model designed for chest X-ray report generation with explicit temporal modelling. Its architecture combines a frozen Rad-DINO~\citep{P_rez_Garc_a_2025} visual encoder with Meditron-7B~\citep{chen2023meditron70b}, connected through a Temporal Alignment Connector. In this work, we use Libra as a pre-trained radiology MLLM backbone and provide only the current frontal-view image as input for consistency with the other models.

\section{Other Experiments}
\label{app:other_exp}

\subsection{Effect of Rollout Size under Beam Search}

\begin{figure*}[ht]
  \centering
  \includegraphics[width=\linewidth]{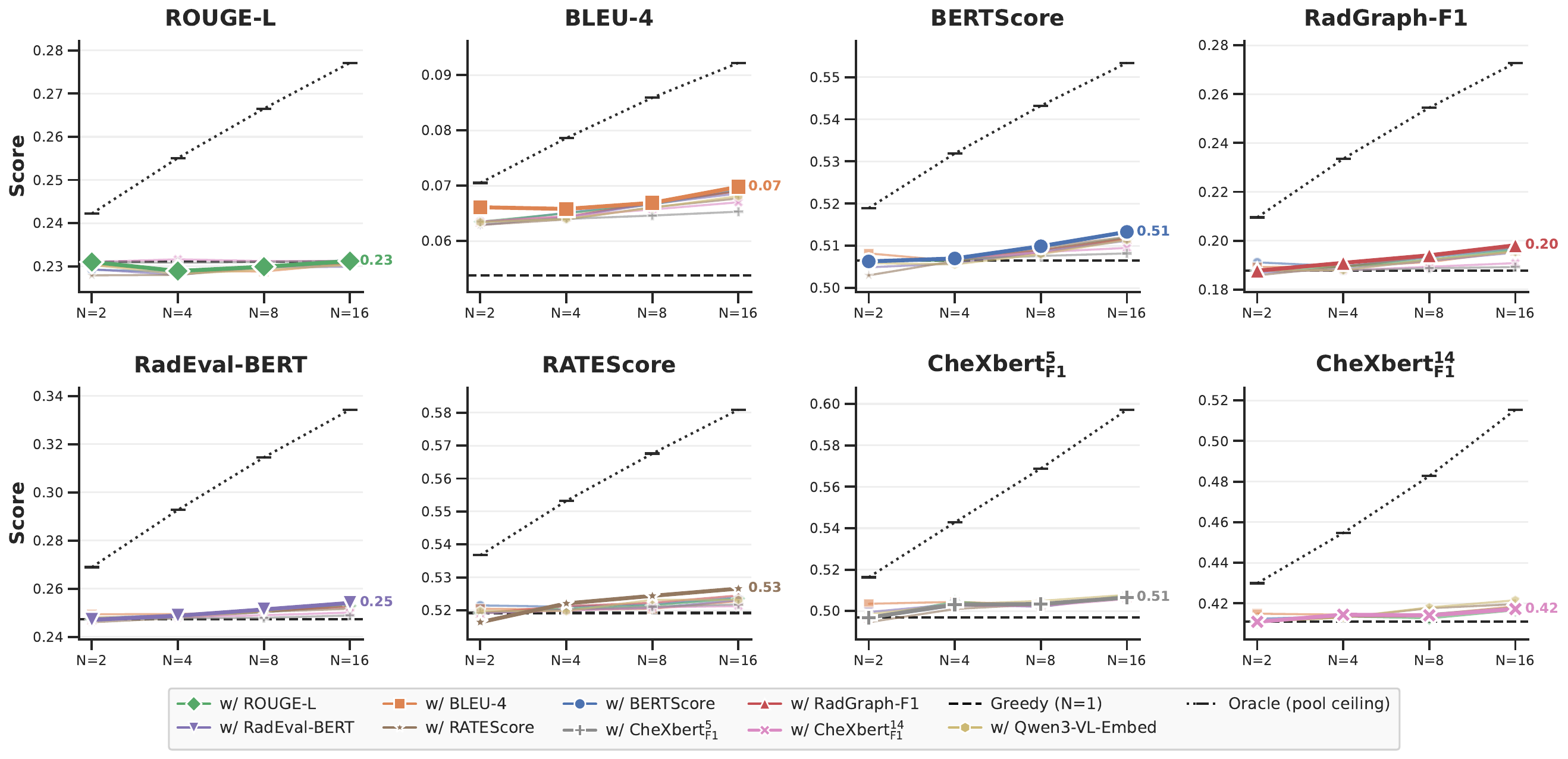}
  \vspace{-23pt}
    \caption{\textbf{Effect of rollout size under different utilities with beam search.}
    Each subplot reports one metric as the beam-search rollout size varies over $N{\in}\{2,4,8,16\}$ under different consensus utilities.}
  \label{fig:metric_trend_beam}
\end{figure*}

Figure~\ref{fig:metric_trend_beam} analyses how rollout size affects selection performance under different consensus utilities when beam search is used for candidate generation. Overall, increasing the rollout size generally improves performance, suggesting that larger candidate pools provide more opportunities for consensus-based selection to recover higher-quality reports. However, gains gradually diminish as $N$ increases, indicating that candidate diversity saturates beyond a certain budget.

Compared with stochastic sampling in Figure~\ref{fig:metric_trend_sampling}, beam search explores candidates in a more likelihood-concentrated manner and typically produces less diverse rollout pools. This is reflected by the lower oracle curves under beam search, which suggest a smaller pool-bounded upper bound than sampling. Nevertheless, \textsc{CCS} still benefits from larger beam-search pools, although the magnitude of improvement varies across utilities. These results indicate that the gains do not rely solely on stochastic exploration, but also arise from more effective candidate selection at inference time.

The oracle curves further reveal a persistent gap between achievable pool quality and actual selection performance, suggesting additional headroom for improving utility design without changing the underlying generator.

\subsection{Effect of Sampling Temperature}

\input{tables/temperature}

Table~\ref{tab:temperature} studies the effect of sampling temperature $\tau$ on candidate generation quality for \textsc{CCS}. Lower temperatures produce more deterministic reports with reduced candidate diversity, whereas higher temperatures increase exploration but may introduce unstable or clinically inconsistent generations.

We observe that moderate sampling temperatures ($\tau \in [0.25,0.50]$) provide the most favourable trade-off between diversity and report quality, yielding consistently strong performance across both lexical and radiology-specific metrics. In contrast, fully deterministic decoding ($\tau=0$) limits the potential of candidate selection, while overly aggressive sampling ($\tau\ge0.75$) reduces overall utility due to noisier candidate pools. Based on these observations, we adopt $\tau=0.5$ as the default setting throughout the paper.

\section{Additional Statement}

Generative AI tools were used only for presentation-level assistance in this work. Specifically, they assisted with colour refinement and visual polishing of the icons in Figure~\ref{fig:case-problem} and Figure~\ref{fig:framework}, with the sole purpose of improving figure readability. These tools were not used to generate scientific claims, conduct analysis, design experiments, or produce results. We also used Overleaf's AI assistant for minor spelling and grammar checks under UK English conventions.

%% file: tables/temperature.tex
\begin{table*}[ht]
\begin{center}
\small
\setlength{\tabcolsep}{2pt}
\resizebox{\textwidth}{!}{%
    \begin{tabular}{l|ccc|c
    c
    c
    c
    c}
        \toprule
        \multirow{3}{*}{\makecell{$\boldsymbol{\tau}$}} & \multicolumn{3}{c|}{\textbf{Lexical Metric}} & \multicolumn{5}{c}{\textbf{Radiology-specific Metric}} \\
        \cmidrule(lr){2-4} \cmidrule(lr){5-9}
        & {ROUGE-L} & {BLEU} & {BERTScore} & {RadGraph-F1} & {RaTEScore} & {RadEval-BERT} & {$\text{CheXbert}_{\text{F1}}^{\text{5}}$} & {$\text{CheXbert}_{\text{F1}}^{\text{14}}$} \\
        \midrule
          \rowcolor{gray!18} \textsc{0.00} & {\bfseries 0.2310} & {\underline{0.0538}} & 0.5065 & {\underline{0.1877}} & {\underline{0.5192}} & 0.2473 & 0.4968 & 0.4109 \\
          \textsc{0.25} & {\underline{0.2299}} & {\bfseries 0.0548} & {\bfseries 0.5163} & 0.1977 & {\bfseries 0.5200} & {\bfseries 0.2505} & {\underline{0.4972}} & 0.4457 \\
          \rowcolor{gray!18} \textsc{0.50}  & 0.2252 & 0.0534 & {\underline{0.5128}} & {\bfseries 0.1989} & 0.5165 & {\underline{0.2493}} & {\bfseries 0.5041} & {\bfseries 0.4519} \\
          \textsc{0.75} & 0.2102 & 0.0482 & 0.5013 & 0.1855 & 0.5086 & 0.2432 & 0.4927 & {\underline{0.4518}} \\
          \rowcolor{gray!18} \textsc{1.00}  & 0.1907 & 0.0427 & 0.4831 & 0.1667 & 0.4943 & 0.2468 & 0.4870 & 0.4416 \\
        \bottomrule
    \end{tabular}
}
\end{center}
\vspace{-12pt}
\caption{\textbf{Ablation study of sampling temperature ($\boldsymbol{\tau}$).}
Effect of sampling temperature on candidate generation for clinical consensus selection, where $\tau=0$ denotes greedy decoding.
\textbf{Best} and \underline{second-best} results are bolded and underlined, respectively.
$\tau \in \{0, 0.25, 0.5, 0.75, 1.0\}$.}
\label{tab:temperature}
\vspace{-10pt}
\end{table*}

%% file: custom.bib
@inproceedings{
li2023llavamed,
title={{LL}a{VA}-Med: Training a Large Language-and-Vision Assistant for Biomedicine in One Day},
author={Chunyuan Li and Cliff Wong and Sheng Zhang and Naoto Usuyama and Haotian Liu and Jianwei Yang and Tristan Naumann and Hoifung Poon and Jianfeng Gao},
booktitle={Thirty-seventh Conference on Neural Information Processing Systems Datasets and Benchmarks Track},
year={2023},
url={https://openreview.net/forum?id=GSuP99u2kR}
}

@misc{bannur2024maira2groundedradiologyreport,
      title={MAIRA-2: Grounded Radiology Report Generation}, 
      author={Shruthi Bannur and Kenza Bouzid and Daniel C. Castro and Anton Schwaighofer and Anja Thieme and Sam Bond-Taylor and Maximilian Ilse and Fernando Pérez-García and Valentina Salvatelli and Harshita Sharma and Felix Meissen and Mercy Ranjit and Shaury Srivastav and Julia Gong and Noel C. F. Codella and Fabian Falck and Ozan Oktay and Matthew P. Lungren and Maria Teodora Wetscherek and Javier Alvarez-Valle and Stephanie L. Hyland},
      year={2024},
      eprint={2406.04449},
      archivePrefix={arXiv},
      primaryClass={cs.CL},
      url={https://arxiv.org/abs/2406.04449}, 
}

@misc{openai2024gpt4technicalreport,
      title={GPT-4 Technical Report}, 
      author={OpenAI and Josh Achiam and Steven Adler and Sandhini Agarwal and Lama Ahmad and Ilge Akkaya and Florencia Leoni Aleman and Diogo Almeida and Janko Altenschmidt and Sam Altman and Shyamal Anadkat and Red Avila and Igor Babuschkin and Suchir Balaji and Valerie Balcom and Paul Baltescu and Haiming Bao and Mohammad Bavarian and Jeff Belgum and Irwan Bello and Jake Berdine and Gabriel Bernadett-Shapiro and Christopher Berner and Lenny Bogdonoff and Oleg Boiko and Madelaine Boyd and Anna-Luisa Brakman and Greg Brockman and Tim Brooks and Miles Brundage and Kevin Button and Trevor Cai and Rosie Campbell and Andrew Cann and Brittany Carey and Chelsea Carlson and Rory Carmichael and Brooke Chan and Che Chang and Fotis Chantzis and Derek Chen and Sully Chen and Ruby Chen and Jason Chen and Mark Chen and Ben Chess and Chester Cho and Casey Chu and Hyung Won Chung and Dave Cummings and Jeremiah Currier and Yunxing Dai and Cory Decareaux and Thomas Degry and Noah Deutsch and Damien Deville and Arka Dhar and David Dohan and Steve Dowling and Sheila Dunning and Adrien Ecoffet and Atty Eleti and Tyna Eloundou and David Farhi and Liam Fedus and Niko Felix and Simón Posada Fishman and Juston Forte and Isabella Fulford and Leo Gao and Elie Georges and Christian Gibson and Vik Goel and Tarun Gogineni and Gabriel Goh and Rapha Gontijo-Lopes and Jonathan Gordon and Morgan Grafstein and Scott Gray and Ryan Greene and Joshua Gross and Shixiang Shane Gu and Yufei Guo and Chris Hallacy and Jesse Han and Jeff Harris and Yuchen He and Mike Heaton and Johannes Heidecke and Chris Hesse and Alan Hickey and Wade Hickey and Peter Hoeschele and Brandon Houghton and Kenny Hsu and Shengli Hu and Xin Hu and Joost Huizinga and Shantanu Jain and Shawn Jain and Joanne Jang and Angela Jiang and Roger Jiang and Haozhun Jin and Denny Jin and Shino Jomoto and Billie Jonn and Heewoo Jun and Tomer Kaftan and Łukasz Kaiser and Ali Kamali and Ingmar Kanitscheider and Nitish Shirish Keskar and Tabarak Khan and Logan Kilpatrick and Jong Wook Kim and Christina Kim and Yongjik Kim and Jan Hendrik Kirchner and Jamie Kiros and Matt Knight and Daniel Kokotajlo and Łukasz Kondraciuk and Andrew Kondrich and Aris Konstantinidis and Kyle Kosic and Gretchen Krueger and Vishal Kuo and Michael Lampe and Ikai Lan and Teddy Lee and Jan Leike and Jade Leung and Daniel Levy and Chak Ming Li and Rachel Lim and Molly Lin and Stephanie Lin and Mateusz Litwin and Theresa Lopez and Ryan Lowe and Patricia Lue and Anna Makanju and Kim Malfacini and Sam Manning and Todor Markov and Yaniv Markovski and Bianca Martin and Katie Mayer and Andrew Mayne and Bob McGrew and Scott Mayer McKinney and Christine McLeavey and Paul McMillan and Jake McNeil and David Medina and Aalok Mehta and Jacob Menick and Luke Metz and Andrey Mishchenko and Pamela Mishkin and Vinnie Monaco and Evan Morikawa and Daniel Mossing and Tong Mu and Mira Murati and Oleg Murk and David Mély and Ashvin Nair and Reiichiro Nakano and Rajeev Nayak and Arvind Neelakantan and Richard Ngo and Hyeonwoo Noh and Long Ouyang and Cullen O'Keefe and Jakub Pachocki and Alex Paino and Joe Palermo and Ashley Pantuliano and Giambattista Parascandolo and Joel Parish and Emy Parparita and Alex Passos and Mikhail Pavlov and Andrew Peng and Adam Perelman and Filipe de Avila Belbute Peres and Michael Petrov and Henrique Ponde de Oliveira Pinto and Michael and Pokorny and Michelle Pokrass and Vitchyr H. Pong and Tolly Powell and Alethea Power and Boris Power and Elizabeth Proehl and Raul Puri and Alec Radford and Jack Rae and Aditya Ramesh and Cameron Raymond and Francis Real and Kendra Rimbach and Carl Ross and Bob Rotsted and Henri Roussez and Nick Ryder and Mario Saltarelli and Ted Sanders and Shibani Santurkar and Girish Sastry and Heather Schmidt and David Schnurr and John Schulman and Daniel Selsam and Kyla Sheppard and Toki Sherbakov and Jessica Shieh and Sarah Shoker and Pranav Shyam and Szymon Sidor and Eric Sigler and Maddie Simens and Jordan Sitkin and Katarina Slama and Ian Sohl and Benjamin Sokolowsky and Yang Song and Natalie Staudacher and Felipe Petroski Such and Natalie Summers and Ilya Sutskever and Jie Tang and Nikolas Tezak and Madeleine B. Thompson and Phil Tillet and Amin Tootoonchian and Elizabeth Tseng and Preston Tuggle and Nick Turley and Jerry Tworek and Juan Felipe Cerón Uribe and Andrea Vallone and Arun Vijayvergiya and Chelsea Voss and Carroll Wainwright and Justin Jay Wang and Alvin Wang and Ben Wang and Jonathan Ward and Jason Wei and CJ Weinmann and Akila Welihinda and Peter Welinder and Jiayi Weng and Lilian Weng and Matt Wiethoff and Dave Willner and Clemens Winter and Samuel Wolrich and Hannah Wong and Lauren Workman and Sherwin Wu and Jeff Wu and Michael Wu and Kai Xiao and Tao Xu and Sarah Yoo and Kevin Yu and Qiming Yuan and Wojciech Zaremba and Rowan Zellers and Chong Zhang and Marvin Zhang and Shengjia Zhao and Tianhao Zheng and Juntang Zhuang and William Zhuk and Barret Zoph},
      year={2024},
      eprint={2303.08774},
      archivePrefix={arXiv},
      primaryClass={cs.CL},
      url={https://arxiv.org/abs/2303.08774}, 
}

@misc{xia2025mmedragversatilemultimodalrag,
      title={MMed-RAG: Versatile Multimodal RAG System for Medical Vision Language Models}, 
      author={Peng Xia and Kangyu Zhu and Haoran Li and Tianze Wang and Weijia Shi and Sheng Wang and Linjun Zhang and James Zou and Huaxiu Yao},
      year={2025},
      eprint={2410.13085},
      archivePrefix={arXiv},
      primaryClass={cs.LG},
      url={https://arxiv.org/abs/2410.13085}, 
}

@misc{hou2025radarenhancingradiologyreport,
      title={RADAR: Enhancing Radiology Report Generation with Supplementary Knowledge Injection}, 
      author={Wenjun Hou and Yi Cheng and Kaishuai Xu and Heng Li and Yan Hu and Wenjie Li and Jiang Liu},
      year={2025},
      eprint={2505.14318},
      archivePrefix={arXiv},
      primaryClass={cs.CV},
      url={https://arxiv.org/abs/2505.14318}, 
}

@misc{liu2023visualinstructiontuning,
      title={Visual Instruction Tuning}, 
      author={Haotian Liu and Chunyuan Li and Qingyang Wu and Yong Jae Lee},
      year={2023},
      eprint={2304.08485},
      archivePrefix={arXiv},
      primaryClass={cs.CV},
      url={https://arxiv.org/abs/2304.08485}, 
}

@inproceedings{zhang2025libraleveragingtemporalimages,
    title = "Libra: Leveraging Temporal Images for Biomedical Radiology Analysis",
    author = "Zhang, Xi  and
      Meng, Zaiqiao  and
      Lever, Jake  and
      Ho, Edmond S. L.",
    editor = "Che, Wanxiang  and
      Nabende, Joyce  and
      Shutova, Ekaterina  and
      Pilehvar, Mohammad Taher",
    booktitle = "Findings of the Association for Computational Linguistics: ACL 2025",
    month = jul,
    year = "2025",
    address = "Vienna, Austria",
    publisher = "Association for Computational Linguistics",
    url = "https://aclanthology.org/2025.findings-acl.888/",
    doi = "10.18653/v1/2025.findings-acl.888",
    pages = "17275--17303",
    ISBN = "979-8-89176-256-5"
}

@misc{tu2023generalistbiomedicalai,
      title={Towards Generalist Biomedical AI}, 
      author={Tao Tu and Shekoofeh Azizi and Danny Driess and Mike Schaekermann and Mohamed Amin and Pi-Chuan Chang and Andrew Carroll and Chuck Lau and Ryutaro Tanno and Ira Ktena and Basil Mustafa and Aakanksha Chowdhery and Yun Liu and Simon Kornblith and David Fleet and Philip Mansfield and Sushant Prakash and Renee Wong and Sunny Virmani and Christopher Semturs and S Sara Mahdavi and Bradley Green and Ewa Dominowska and Blaise Aguera y Arcas and Joelle Barral and Dale Webster and Greg S. Corrado and Yossi Matias and Karan Singhal and Pete Florence and Alan Karthikesalingam and Vivek Natarajan},
      year={2023},
      eprint={2307.14334},
      archivePrefix={arXiv},
      primaryClass={cs.CL},
      url={https://arxiv.org/abs/2307.14334}, 
}

@article{Zambrano_Chaves_2025,
   title={A clinically accessible small multimodal radiology model and evaluation metric for chest X-ray findings},
   volume={16},
   ISSN={2041-1723},
   url={http://dx.doi.org/10.1038/s41467-025-58344-x},
   DOI={10.1038/s41467-025-58344-x},
   number={1},
   journal={Nature Communications},
   publisher={Springer Science and Business Media LLC},
   author={Zambrano Chaves, Juan Manuel and Huang, Shih-Cheng and Xu, Yanbo and Xu, Hanwen and Usuyama, Naoto and Zhang, Sheng and Wang, Fei and Xie, Yujia and Khademi, Mahmoud and Yang, Ziyi and Awadalla, Hany and Gong, Julia and Hu, Houdong and Yang, Jianwei and Li, Chunyuan and Gao, Jianfeng and Gu, Yu and Wong, Cliff and Wei, Mu and Naumann, Tristan and Chen, Muhao and Lungren, Matthew P. and Chaudhari, Akshay and Yeung-Levy, Serena and Langlotz, Curtis P. and Wang, Sheng and Poon, Hoifung},
   year={2025},
   month=apr }

@misc{hyland2024maira1specialisedlargemultimodal,
      title={MAIRA-1: A specialised large multimodal model for radiology report generation}, 
      author={Stephanie L. Hyland and Shruthi Bannur and Kenza Bouzid and Daniel C. Castro and Mercy Ranjit and Anton Schwaighofer and Fernando Pérez-García and Valentina Salvatelli and Shaury Srivastav and Anja Thieme and Noel Codella and Matthew P. Lungren and Maria Teodora Wetscherek and Ozan Oktay and Javier Alvarez-Valle},
      year={2024},
      eprint={2311.13668},
      archivePrefix={arXiv},
      primaryClass={cs.CL},
      url={https://arxiv.org/abs/2311.13668}, 
}

@article{johnson2019mimic,
  title={MIMIC-CXR, a de-identified publicly available database of chest radiographs with free-text reports},
  author={Johnson, Alistair EW and Pollard, Tom J and Berkowitz, Seth J and Greenbaum, Nathaniel R and Lungren, Matthew P and Deng, Chih-ying and Mark, Roger G and Horng, Steven},
  journal={Scientific data},
  volume={6},
  number={1},
  pages={317},
  year={2019},
  publisher={Nature Publishing Group UK London},
  doi={https://doi.org/10.1038/s41597-019-0322-0}
}

@inproceedings{lin-2004-rouge,
    title = "{ROUGE}: A Package for Automatic Evaluation of Summaries",
    author = "Lin, Chin-Yew",
    booktitle = "Text Summarization Branches Out",
    month = jul,
    year = "2004",
    address = "Barcelona, Spain",
    publisher = "Association for Computational Linguistics",
    url = "https://aclanthology.org/W04-1013/",
    pages = "74--81"
}

@inproceedings{10.3115/1073083.1073135,
author = {Papineni, Kishore and Roukos, Salim and Ward, Todd and Zhu, Wei-Jing},
title = {BLEU: a method for automatic evaluation of machine translation},
year = {2002},
publisher = {Association for Computational Linguistics},
address = {USA},
url = {https://doi.org/10.3115/1073083.1073135},
doi = {10.3115/1073083.1073135},
booktitle = {Proceedings of the 40th Annual Meeting on Association for Computational Linguistics},
pages = {311–318},
numpages = {8},
location = {Philadelphia, Pennsylvania},
series = {ACL '02}
}

@misc{zhang2020bertscoreevaluatingtextgeneration,
      title={BERTScore: Evaluating Text Generation with BERT}, 
      author={Tianyi Zhang and Varsha Kishore and Felix Wu and Kilian Q. Weinberger and Yoav Artzi},
      year={2020},
      eprint={1904.09675},
      archivePrefix={arXiv},
      primaryClass={cs.CL},
      url={https://arxiv.org/abs/1904.09675}, 
}

@inproceedings{delbrouck-etal-2022-improving,
    title = "Improving the Factual Correctness of Radiology Report Generation with Semantic Rewards",
    author = "Delbrouck, Jean-Benoit  and
      Chambon, Pierre  and
      Bluethgen, Christian  and
      Tsai, Emily  and
      Almusa, Omar  and
      Langlotz, Curtis",
    editor = "Goldberg, Yoav  and
      Kozareva, Zornitsa  and
      Zhang, Yue",
    booktitle = "Findings of the Association for Computational Linguistics: EMNLP 2022",
    month = dec,
    year = "2022",
    address = "Abu Dhabi, United Arab Emirates",
    publisher = "Association for Computational Linguistics",
    url = "https://aclanthology.org/2022.findings-emnlp.319/",
    doi = "10.18653/v1/2022.findings-emnlp.319",
    pages = "4348--4360"
}

@inproceedings{zhao2024ratescoremetricradiologyreport,
    title = "{R}a{TES}core: A Metric for Radiology Report Generation",
    author = "Zhao, Weike  and
      Wu, Chaoyi  and
      Zhang, Xiaoman  and
      Zhang, Ya  and
      Wang, Yanfeng  and
      Xie, Weidi",
    editor = "Al-Onaizan, Yaser  and
      Bansal, Mohit  and
      Chen, Yun-Nung",
    booktitle = "Proceedings of the 2024 Conference on Empirical Methods in Natural Language Processing",
    month = nov,
    year = "2024",
    address = "Miami, Florida, USA",
    publisher = "Association for Computational Linguistics",
    url = "https://aclanthology.org/2024.emnlp-main.836/",
    doi = "10.18653/v1/2024.emnlp-main.836",
    pages = "15004--15019"
}

@misc{smit2020chexbertcombiningautomaticlabelers,
      title={CheXbert: Combining Automatic Labelers and Expert Annotations for Accurate Radiology Report Labeling Using BERT}, 
      author={Akshay Smit and Saahil Jain and Pranav Rajpurkar and Anuj Pareek and Andrew Y. Ng and Matthew P. Lungren},
      year={2020},
      eprint={2004.09167},
      archivePrefix={arXiv},
      primaryClass={cs.CL},
      url={https://arxiv.org/abs/2004.09167}, 
}

@misc{irvin2019chexpertlargechestradiograph,
      title={CheXpert: A Large Chest Radiograph Dataset with Uncertainty Labels and Expert Comparison}, 
      author={Jeremy Irvin and Pranav Rajpurkar and Michael Ko and Yifan Yu and Silviana Ciurea-Ilcus and Chris Chute and Henrik Marklund and Behzad Haghgoo and Robyn Ball and Katie Shpanskaya and Jayne Seekins and David A. Mong and Safwan S. Halabi and Jesse K. Sandberg and Ricky Jones and David B. Larson and Curtis P. Langlotz and Bhavik N. Patel and Matthew P. Lungren and Andrew Y. Ng},
      year={2019},
      eprint={1901.07031},
      archivePrefix={arXiv},
      primaryClass={cs.CV},
      url={https://arxiv.org/abs/1901.07031}, 
}

@article{johnson2018mimic,
  title={The MIMIC Code Repository: enabling reproducibility in critical care research},
  author={Johnson, Alistair EW and Stone, David J and Celi, Leo A and Pollard, Tom J},
  journal={Journal of the American Medical Informatics Association},
  volume={25},
  number={1},
  pages={32--39},
  year={2018},
  publisher={Oxford Academic},
  doi={https://doi.org/10.1093/jamia/ocx084}
}

@misc{li2023contrastivedecodingopenendedtext,
      title={Contrastive Decoding: Open-ended Text Generation as Optimization}, 
      author={Xiang Lisa Li and Ari Holtzman and Daniel Fried and Percy Liang and Jason Eisner and Tatsunori Hashimoto and Luke Zettlemoyer and Mike Lewis},
      year={2023},
      eprint={2210.15097},
      archivePrefix={arXiv},
      primaryClass={cs.CL},
      url={https://arxiv.org/abs/2210.15097}, 
}

@article{demner2015preparing,
  title={Preparing a collection of radiology examinations for distribution and retrieval},
  author={Demner-Fushman, Dina and Kohli, Marc D and Rosenman, Marc B and Shooshan, Sonya E and Rodriguez, Laritza and Antani, Sameer and Thoma, George R and McDonald, Clement J},
  journal={Journal of the American Medical Informatics Association},
  volume={23},
  number={2},
  pages={304--310},
  year={2015},
  publisher={Oxford Academic},
  doi={https://doi.org/10.1093/jamia/ocv080}
}

@misc{chambon2024chexpert,
      title={CheXpert Plus: Augmenting a Large Chest X-ray Dataset with Text Radiology Reports, Patient Demographics and Additional Image Formats}, 
      author={Pierre Chambon and Jean-Benoit Delbrouck and Thomas Sounack and Shih-Cheng Huang and Zhihong Chen and Maya Varma and Steven QH Truong and Chu The Chuong and Curtis P. Langlotz},
      year={2024},
      eprint={2405.19538},
      archivePrefix={arXiv},
      primaryClass={cs.CL},
      url={https://arxiv.org/abs/2405.19538}, 
}

@misc{johnson2019mimiccxrjpglargepubliclyavailable,
      title={MIMIC-CXR-JPG, a large publicly available database of labeled chest radiographs}, 
      author={Alistair E. W. Johnson and Tom J. Pollard and Nathaniel R. Greenbaum and Matthew P. Lungren and Chih-ying Deng and Yifan Peng and Zhiyong Lu and Roger G. Mark and Seth J. Berkowitz and Steven Horng},
      year={2019},
      eprint={1901.07042},
      archivePrefix={arXiv},
      primaryClass={cs.CV},
      url={https://arxiv.org/abs/1901.07042}, 
}

@misc{devlin2019bertpretrainingdeepbidirectional,
      title={BERT: Pre-training of Deep Bidirectional Transformers for Language Understanding}, 
      author={Jacob Devlin and Ming-Wei Chang and Kenton Lee and Kristina Toutanova},
      year={2019},
      eprint={1810.04805},
      archivePrefix={arXiv},
      primaryClass={cs.CL},
      url={https://arxiv.org/abs/1810.04805}, 
}

@misc{warner2024smarterbetterfasterlonger,
      title={Smarter, Better, Faster, Longer: A Modern Bidirectional Encoder for Fast, Memory Efficient, and Long Context Finetuning and Inference}, 
      author={Benjamin Warner and Antoine Chaffin and Benjamin Clavié and Orion Weller and Oskar Hallström and Said Taghadouini and Alexis Gallagher and Raja Biswas and Faisal Ladhak and Tom Aarsen and Nathan Cooper and Griffin Adams and Jeremy Howard and Iacopo Poli},
      year={2024},
      eprint={2412.13663},
      archivePrefix={arXiv},
      primaryClass={cs.CL},
      url={https://arxiv.org/abs/2412.13663}, 
}

@misc{perezgarcia2024raddino,
      title={{RAD-DINO}: Exploring Scalable Medical Image Encoders Beyond Text Supervision},
      author={Fernando Pérez-García and Harshita Sharma and Sam Bond-Taylor and Kenza Bouzid and Valentina Salvatelli and Maximilian Ilse and Shruthi Bannur and Daniel C. Castro and Anton Schwaighofer and Matthew P. Lungren and Maria Wetscherek and Noel Codella and Stephanie L. Hyland and Javier Alvarez-Valle and Ozan Oktay},
      year={2024},
      eprint={2401.10815},
      archivePrefix={arXiv},
      primaryClass={cs.CV}
}

@misc{vicuna2023,
    title = {Vicuna: An Open-Source Chatbot Impressing GPT-4 with 90\%* ChatGPT Quality},
    url = {https://lmsys.org/blog/2023-03-30-vicuna/},
    author = {Chiang, Wei-Lin and Li, Zhuohan and Lin, Zi and Sheng, Ying and Wu, Zhanghao and Zhang, Hao and Zheng, Lianmin and Zhuang, Siyuan and Zhuang, Yonghao and Gonzalez, Joseph E. and Stoica, Ion and Xing, Eric P.},
    month = {March},
    year = {2023}
}

@article{P_rez_Garc_a_2025,
   title={Exploring scalable medical image encoders beyond text supervision},
   volume={7},
   ISSN={2522-5839},
   url={http://dx.doi.org/10.1038/s42256-024-00965-w},
   DOI={10.1038/s42256-024-00965-w},
   number={1},
   journal={Nature Machine Intelligence},
   publisher={Springer Science and Business Media LLC},
   author={Pérez-García, Fernando and Sharma, Harshita and Bond-Taylor, Sam and Bouzid, Kenza and Salvatelli, Valentina and Ilse, Maximilian and Bannur, Shruthi and Castro, Daniel C. and Schwaighofer, Anton and Lungren, Matthew P. and Wetscherek, Maria Teodora and Codella, Noel and Hyland, Stephanie L. and Alvarez-Valle, Javier and Oktay, Ozan},
   year={2025},
   month=jan, pages={119–130} }

@misc{chen2023meditron70b,
      title={MEDITRON-70B: Scaling Medical Pretraining for Large Language Models}, 
      author={Zeming Chen and Alejandro Hernández-Cano and Angelika Romanou and Antoine Bonnet and Kyle Matoba and Francesco Salvi and Matteo Pagliardini and Simin Fan and Andreas Köpf and Amirkeivan Mohtashami and Alexandre Sallinen and Alireza Sakhaeirad and Vinitra Swamy and Igor Krawczuk and Deniz Bayazit and Axel Marmet and Syrielle Montariol and Mary-Anne Hartley and Martin Jaggi and Antoine Bosselut},
      year={2023},
      eprint={2311.16079},
      archivePrefix={arXiv},
      primaryClass={cs.CL}
}

@misc{hu2021loralowrankadaptationlarge,
      title={LoRA: Low-Rank Adaptation of Large Language Models}, 
      author={Edward J. Hu and Yelong Shen and Phillip Wallis and Zeyuan Allen-Zhu and Yuanzhi Li and Shean Wang and Lu Wang and Weizhu Chen},
      year={2021},
      eprint={2106.09685},
      archivePrefix={arXiv},
      primaryClass={cs.CL},
      url={https://arxiv.org/abs/2106.09685}, 
}

@misc{zhang2025biomedclipmultimodalbiomedicalfoundation,
      title={BiomedCLIP: a multimodal biomedical foundation model pretrained from fifteen million scientific image-text pairs}, 
      author={Sheng Zhang and Yanbo Xu and Naoto Usuyama and Hanwen Xu and Jaspreet Bagga and Robert Tinn and Sam Preston and Rajesh Rao and Mu Wei and Naveen Valluri and Cliff Wong and Andrea Tupini and Yu Wang and Matt Mazzola and Swadheen Shukla and Lars Liden and Jianfeng Gao and Angela Crabtree and Brian Piening and Carlo Bifulco and Matthew P. Lungren and Tristan Naumann and Sheng Wang and Hoifung Poon},
      year={2025},
      eprint={2303.00915},
      archivePrefix={arXiv},
      primaryClass={cs.CV},
      url={https://arxiv.org/abs/2303.00915}, 
}

@misc{jiang2023mistral7b,
      title={Mistral 7B}, 
      author={Albert Q. Jiang and Alexandre Sablayrolles and Arthur Mensch and Chris Bamford and Devendra Singh Chaplot and Diego de las Casas and Florian Bressand and Gianna Lengyel and Guillaume Lample and Lucile Saulnier and Lélio Renard Lavaud and Marie-Anne Lachaux and Pierre Stock and Teven Le Scao and Thibaut Lavril and Thomas Wang and Timothée Lacroix and William El Sayed},
      year={2023},
      eprint={2310.06825},
      archivePrefix={arXiv},
      primaryClass={cs.CL},
      url={https://arxiv.org/abs/2310.06825}, 
}

@misc{radford2021learningtransferablevisualmodels,
      title={Learning Transferable Visual Models From Natural Language Supervision}, 
      author={Alec Radford and Jong Wook Kim and Chris Hallacy and Aditya Ramesh and Gabriel Goh and Sandhini Agarwal and Girish Sastry and Amanda Askell and Pamela Mishkin and Jack Clark and Gretchen Krueger and Ilya Sutskever},
      year={2021},
      eprint={2103.00020},
      archivePrefix={arXiv},
      primaryClass={cs.CV},
      url={https://arxiv.org/abs/2103.00020}, 
}

@misc{zhai2023sigmoidlosslanguageimage,
      title={Sigmoid Loss for Language Image Pre-Training}, 
      author={Xiaohua Zhai and Basil Mustafa and Alexander Kolesnikov and Lucas Beyer},
      year={2023},
      eprint={2303.15343},
      archivePrefix={arXiv},
      primaryClass={cs.CV},
      url={https://arxiv.org/abs/2303.15343}, 
}

@misc{liu2019clinicallyaccuratechestxray,
      title={Clinically Accurate Chest X-Ray Report Generation}, 
      author={Guanxiong Liu and Tzu-Ming Harry Hsu and Matthew McDermott and Willie Boag and Wei-Hung Weng and Peter Szolovits and Marzyeh Ghassemi},
      year={2019},
      eprint={1904.02633},
      archivePrefix={arXiv},
      primaryClass={cs.CV},
      url={https://arxiv.org/abs/1904.02633}, 
}

@article{monshi2020deep,
  title={Deep learning in generating radiology reports: A survey},
  author={Monshi, Maram Mahmoud A and Poon, Josiah and Chung, Vera},
  journal={Artificial Intelligence in Medicine},
  volume={106},
  pages={101878},
  year={2020},
  doi={https://doi.org/10.1016/j.artmed.2020.101878}
}

@misc{sun2025factawaremultimodalretrievalaugmentation,
      title={Fact-Aware Multimodal Retrieval Augmentation for Accurate Medical Radiology Report Generation}, 
      author={Liwen Sun and James Zhao and Megan Han and Chenyan Xiong},
      year={2025},
      eprint={2407.15268},
      archivePrefix={arXiv},
      primaryClass={cs.CL},
      url={https://arxiv.org/abs/2407.15268}, 
}

@misc{wang2018tienettextimageembeddingnetwork,
      title={TieNet: Text-Image Embedding Network for Common Thorax Disease Classification and Reporting in Chest X-rays}, 
      author={Xiaosong Wang and Yifan Peng and Le Lu and Zhiyong Lu and Ronald M. Summers},
      year={2018},
      eprint={1801.04334},
      archivePrefix={arXiv},
      primaryClass={cs.CV},
      url={https://arxiv.org/abs/1801.04334}, 
}

@inproceedings{xu-etal-2025-radeval,
    title = "{R}ad{E}val: A framework for radiology text evaluation",
    author = "Xu, Justin  and
      Zhang, Xi  and
      Abderezaei, Javid  and
      Bauml, Julie  and
      Boodoo, Roger  and
      Haghighi, Fatemeh  and
      Ganjizadeh, Ali  and
      Brattain, Eric  and
      Van Veen, Dave  and
      Meng, Zaiqiao  and
      Eyre, David W  and
      Delbrouck, Jean-Benoit",
    editor = {Habernal, Ivan  and
      Schulam, Peter  and
      Tiedemann, J{\"o}rg},
    booktitle = "Proceedings of the 2025 Conference on Empirical Methods in Natural Language Processing: System Demonstrations",
    month = nov,
    year = "2025",
    address = "Suzhou, China",
    publisher = "Association for Computational Linguistics",
    url = "https://aclanthology.org/2025.emnlp-demos.40/",
    doi = "10.18653/v1/2025.emnlp-demos.40",
    pages = "546--557",
    ISBN = "979-8-89176-334-0"
}

@article{zhang2025automated,
  title        = {Automated chest X-ray report generation remains unsolved},
  author       = {Zhang, Xiaoman and Acosta, Julian Nicolas and Yang, Xiaoli and Adithan, Subathra and Luo, Luyang and Zhou, Hong-Yu and Miller, Joshua and Huang, Ouwen and Zhou, Zongwei and Hamamci, Ibrahim Ethem and Bannur, Shruthi and Bouzid, Kenza and Zhang, Xi and Meng, Zaiqiao and Nicolson, Aaron and Koopman, Bevan and Baek, Inhyeok and Ko, Hanbin and Ranjit, Mercy Prasanna and Srivastav, Shaury and Sambanthan, Sriram Gnana and Rajpurkar, Pranav},
  booktitle={Biocomputing 2026: Proceedings of the Pacific Symposium},
  journal = {Biocomputing 2026: Proceedings of the Pacific Symposium},
  pages={236--250},
  year={2025},
  organization={World Scientific},
  doi = {10.1142/9789819824755_0017},
  url          = {https://doi.org/10.7490/f1000research.1120296.1}
}

@article{zhang2025ccd,
  title={CCD: Mitigating Hallucinations in Radiology MLLMs via Clinical Contrastive Decoding},
  author={Zhang, Xi and Meng, Zaiqiao and Lever, Jake and Ho, Edmond SL},
  journal={arXiv preprint arXiv:2509.23379},
  url={https://arxiv.org/abs/2509.23379}, 
  year={2025}
}

@inproceedings{
kang2026scalable,
title={Scalable Best-of-N Selection for Large Language Models via Self-Certainty},
author={Zhewei Kang and Xuandong Zhao and Dawn Song},
booktitle={The Thirty-ninth Annual Conference on Neural Information Processing Systems},
year={2026},
url={https://openreview.net/forum?id=29FRqmVQK8}
}

@article{choi2026modex,
  title={ModeX: Evaluator-Free Best-of-N Selection for Open-Ended Generation},
  author={Choi, Hyeong Kyu and Li, Sharon},
  journal={arXiv preprint arXiv:2601.02535},
  url={https://arxiv.org/abs/2601.02535},
  year={2026}
}

@article{hu2024can,
  title={Can perplexity reflect large language model's ability in long text understanding?},
  author={Hu, Yutong and Huang, Quzhe and Tao, Mingxu and Zhang, Chen and Feng, Yansong},
  journal={arXiv preprint arXiv:2405.06105},
  year={2024},
  url={https://arxiv.org/abs/2405.06105}
}

@article{li2026qwen3,
  title={Qwen3-VL-Embedding and Qwen3-VL-Reranker: A Unified Framework for State-of-the-Art Multimodal Retrieval and Ranking},
  author={Li, Mingxin and Zhang, Yanzhao and Long, Dingkun and Chen, Keqin and Song, Sibo and Bai, Shuai and Yang, Zhibo and Xie, Pengjun and Yang, An and Liu, Dayiheng and others},
  journal={arXiv preprint arXiv:2601.04720},
  url={https://arxiv.org/abs/2601.04720},
  year={2026}
}

@article{meng2025vlm2vecv2advancingmultimodalembedding,
      title={VLM2Vec-V2: Advancing Multimodal Embedding for Videos, Images, and Visual Documents}, 
      author={Rui Meng and Ziyan Jiang and Ye Liu and Mingyi Su and Xinyi Yang and Yuepeng Fu and Can Qin and Zeyuan Chen and Ran Xu and Caiming Xiong and Yingbo Zhou and Wenhu Chen and Semih Yavuz},
      year={2025},
      eprint={2507.04590},
      archivePrefix={arXiv},
      primaryClass={cs.CV},
      journal={https://arxiv.org/abs/2507.04590}, 
}

@article{mchugh2012interrater,
  title={Interrater reliability: the kappa statistic},
  author={McHugh, Mary L},
  journal={Biochemia medica},
  volume={22},
  number={3},
  pages={276--282},
  year={2012},
  doi={https://pmc.ncbi.nlm.nih.gov/articles/PMC3900052/},
  publisher={Hrvatsko dru{\v{s}}tvo za medicinsku biokemiju i laboratorijsku medicinu}
}

@article{landis1977measurement,
  title={The measurement of observer agreement for categorical data},
  author={Landis, J Richard and Koch, Gary G},
  journal={biometrics},
  pages={159--174},
  year={1977},
  doi={https://doi.org/10.2307/2529310},
  publisher={JSTOR}
}

@inproceedings{
snell2024scaling,
title={Scaling {LLM} Test-Time Compute Optimally Can be More Effective than Scaling Parameters for Reasoning},
author={Charlie Victor Snell and Jaehoon Lee and Kelvin Xu and Aviral Kumar},
booktitle={The Thirteenth International Conference on Learning Representations},
year={2025},
url={https://openreview.net/forum?id=4FWAwZtd2n}
}

@inproceedings{wang2024soft,
  title={Soft self-consistency improves language models agents},
  author={Wang, Han and Prasad, Archiki and Stengel-Eskin, Elias and Bansal, Mohit},
  booktitle={Proceedings of the 62nd Annual Meeting of the Association for Computational Linguistics (Volume 2: Short Papers)},
  url = "https://aclanthology.org/2024.acl-short.28/",
  doi = "10.18653/v1/2024.acl-short.28",
  pages={287--301},
  year={2024}
}

@article{huang2025best,
  title={Is best-of-n the best of them? coverage, scaling, and optimality in inference-time alignment},
  author={Huang, Audrey and Block, Adam and Liu, Qinghua and Jiang, Nan and Krishnamurthy, Akshay and Foster, Dylan J},
  journal={arXiv preprint arXiv:2503.21878},
  year={2025},
  url={https://arxiv.org/abs/2503.21878}
}

@article{shao2024deepseekmath,
  title={Deepseekmath: Pushing the limits of mathematical reasoning in open language models, 2024},
  author={Shao, Zhihong and Wang, Peiyi and Zhu, Qihao and Xu, Runxin and Song, Junxiao and Bi, Xiao and Zhang, Haowei and Zhang, Mingchuan and Li, YK and Wu, Yang and others},
  journal={URL https://arxiv. org/abs/2402.03300},
  url={https://arxiv.org/abs/2402.03300},
  volume={2},
  number={3},
  pages={5},
  year={2024}
}
